\documentclass[letterpaper]{article} 
\usepackage[preprint]{aaai2027}
\usepackage[hyphens]{url} 
\usepackage{graphicx} 
\usepackage{natbib} 
\usepackage{caption} 
\usepackage{booktabs}
\usepackage{amsmath,amssymb}
\usepackage{array}
\usepackage{placeins}
\newlength{\sgllteaserheight}
\newcommand{\sg}{\operatorname{sg}}
\newcommand{\mean}{\operatorname{Mean}}
\newcounter{suppalgorithm}
\newenvironment{algorithm}[1][]{%
  \refstepcounter{suppalgorithm}%
  \begin{center}\begin{minipage}{0.96\columnwidth}\hrule\vspace{3pt}%
  \renewcommand{\caption}[1]{\textbf{Algorithm \thesuppalgorithm: ##1}\par\smallskip}%
}{%
  \vspace{3pt}\hrule\end{minipage}\end{center}%
}
\newenvironment{algorithmic}[1][]{%
  \small\begin{enumerate}\setlength{\itemsep}{1pt}\setlength{\parskip}{0pt}%
}{%
  \end{enumerate}%
}
\newcommand{\Require}{\item[\textbf{Input:}]}
\newcommand{\State}{\item}
\newcommand{\For}[1]{\item \textbf{for} #1 \textbf{do}}
\newcommand{\EndFor}{\item[]\textbf{end for}}
\newcommand{\Return}{\textbf{return} }

\title{Spatially Grounded Latent Lookahead for Vision-Language-Action Models}
\author{
    Qihang Chen\textsuperscript{\rm 1},
    Tengjie Li\textsuperscript{\rm 1},
    Hongwei Lu\textsuperscript{\rm 1},
    Jason Wu\textsuperscript{\rm 2},
    Chyna Qu\textsuperscript{\rm 2},
    Shikui Tu\textsuperscript{\rm 1}\thanks{Corresponding authors.},
    Lei Xu\textsuperscript{\rm 1}\footnotemark[1]
}
\affiliations{
    \textsuperscript{\rm 1}School of Computer Science, Shanghai Jiao Tong University\\
    \textsuperscript{\rm 2}PrismaX Robotics
}

\begin{document}

\maketitle

\begin{abstract}

Vision-language-action (VLA) models benefit from strong semantic representations that support object recognition and instruction understanding. However, precise manipulation also depends on resolving object locations and workspace relations, and on selecting actions with an estimate of how the scene will evolve. Existing geometry-aware methods make the current scene more explicit, while predictive methods model future observations or representations. Yet future information is often neither grounded in a control-relevant spatial state nor directly used to guide action generation. We introduce \emph{Spatially Grounded Latent Lookahead} (SG-LL), which uses the grounded present as the reference for predicting an actionable near-future representation. A frozen spatial teacher first distills geometric priors into policy-connected spatial tokens. A current-conditioned bottleneck then learns a compact latent that captures future information relative to the present scene. This latent is predicted from the current policy context and initial Action Expert features, without access to future observations or the spatial teacher. Finally, zero-initialized residual fusion incorporates the predicted latent into the Action Expert to refine action generation. Experiments on LIBERO-Plus, CALVIN, and three real-world manipulation tasks demonstrate improvements over $\pi_{0.5}$ in both simulation and real-robot settings.

\end{abstract}

\begin{links}
    \link{Code}{https://github.com/Cqhddyb/SG-LL}
\end{links}

\section{Introduction}

Vision-language-action (VLA) models provide a unified policy that maps visual observations and language instructions to robot actions. Large-scale pretraining and robot data have produced increasingly capable generalist policies, including the RT, OpenVLA, and \(\pi\) families \citep{brohan2022rt,kim2024openvla,black2024pi0,physicalintelligence2025pi05}. These models execute diverse manipulation instructions and transfer across task
conditions, yet strong task performance leaves open whether their internal representations explicitly expose the scene geometry needed for control and whether action generation anticipates how the interaction is likely to evolve. This distinction matters under variations in viewpoint,
robot state, object configuration, and observation noise, where a semantically
competent policy must still resolve precise spatial relations
\citep{fei2025libero}.

Reliable manipulation calls for both spatial grounding and predictive
lookahead. Semantic visual features can identify an object without making its
location, robot--object relation, or local configuration readily available to
the Action Expert. Conversely, directly mapping the current context to an Action Chunk does not explicitly represent the near-future scene evolution implied by the current task and policy context. This motivates the central
question of this work: \emph{Can a VLA ground the present scene with control-relevant spatial structure, anticipate where the interaction is heading, and use that lookahead to refine control?}

Spatially aware robot policies address the first requirement through depth,
point clouds, explicit 3D representations, or geometry-aligned features
\citep{shridhar2021cliport,pmlr-v205-shridhar23a,ze20243d,li2025spatial,qu2025spatialvla}. Such
representations can make scene structure more accessible, but they are often
introduced through an additional online modality or a specialized spatial
architecture. Our objective is instead to use a frozen visual geometry model solely as a training-time supervisory reference while retaining the standard RGB policy interface.
Its geometric priors are distilled into a policy-connected current-state representation, while the teacher itself is absent from the deployed policy.

Predictive representations address the second requirement from another
direction. Prior work predicts images, videos, latent futures, or world-action
features for representation learning, visual subgoals, action conditioning, or
rollout-based control \citep{du2023learning,black2024zero,zheng2025flare,su2026world,fang2026privileged,zhao2026frappe,zhang2026dreamvla,miao2026jepa}.
These studies establish future-oriented representation learning as a promising direction for robot control. This raises a key interface challenge: how can a compact, future-informative representation be integrated into an existing flow-matching VLA such that prediction directly informs action generation? In SG-LL, lookahead takes the form of a compact near-future latent predicted once from the current policy context and directly used to refine the inherited action computation.

\begin{figure}[t]
\centering
\setlength{\sgllteaserheight}{0.207\columnwidth}
\begin{minipage}[t]{0.315\columnwidth}
\centering
\parbox[c][\sgllteaserheight][c]{\linewidth}{\centering
\includegraphics[height=\sgllteaserheight]{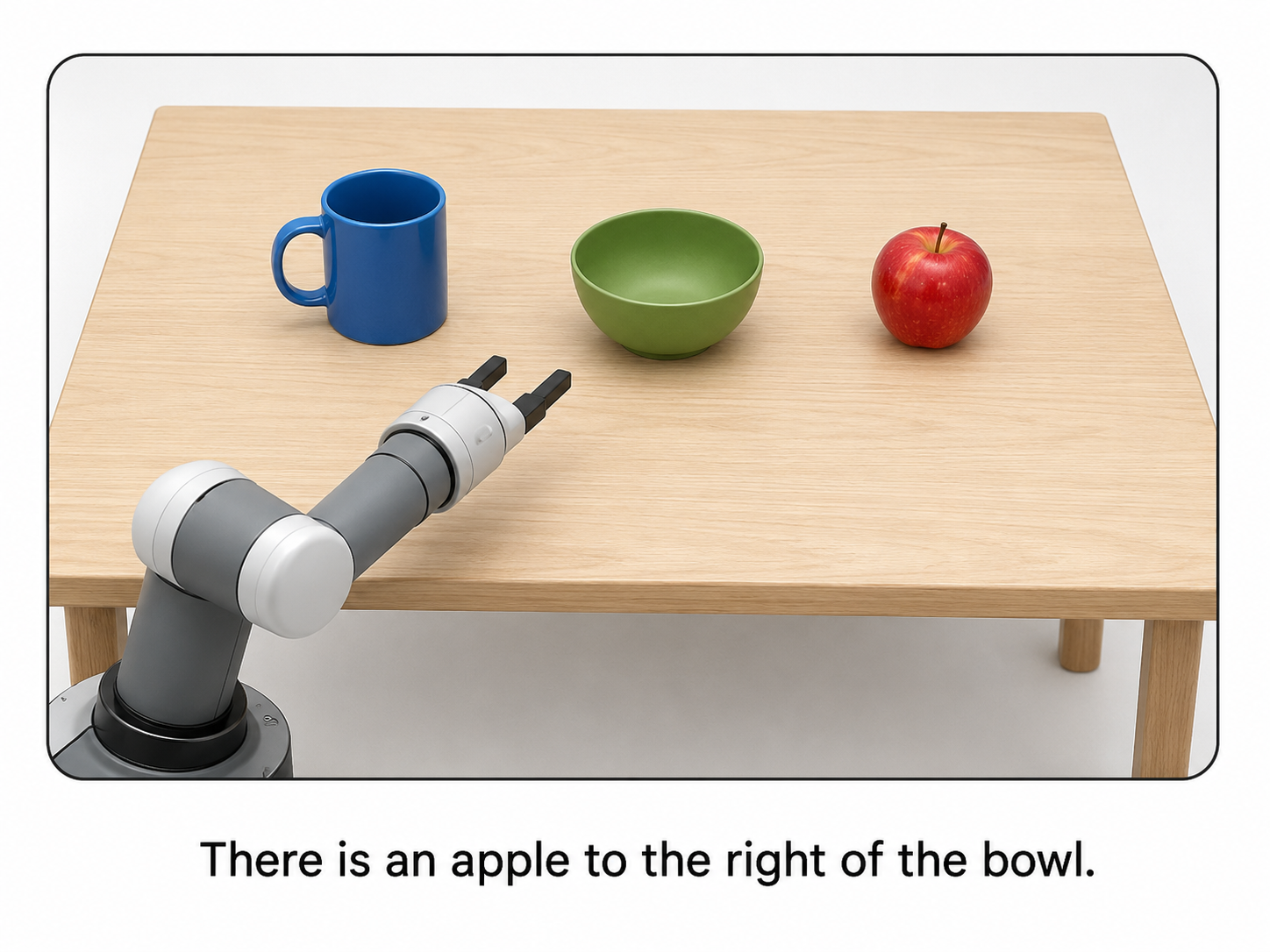}}
\par\small (1) Ground the Present
\end{minipage}\hfill
\begin{minipage}[t]{0.315\columnwidth}
\centering
\parbox[c][\sgllteaserheight][c]{\linewidth}{\centering
\includegraphics[height=\sgllteaserheight]{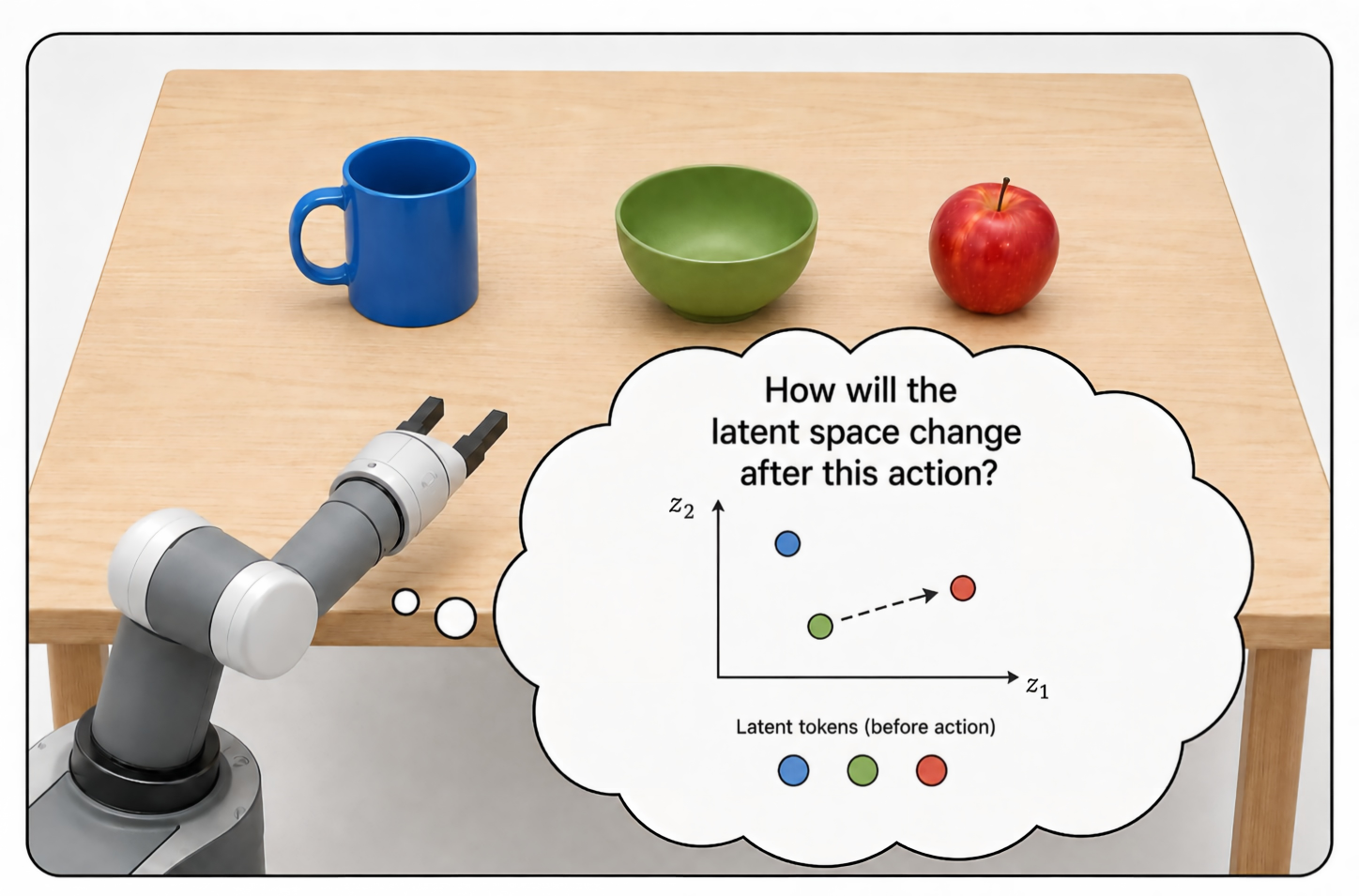}}
\par\small (2) Look Ahead in Latent Space
\end{minipage}\hfill
\begin{minipage}[t]{0.315\columnwidth}
\centering
\parbox[c][\sgllteaserheight][c]{\linewidth}{\centering
\includegraphics[height=\sgllteaserheight]{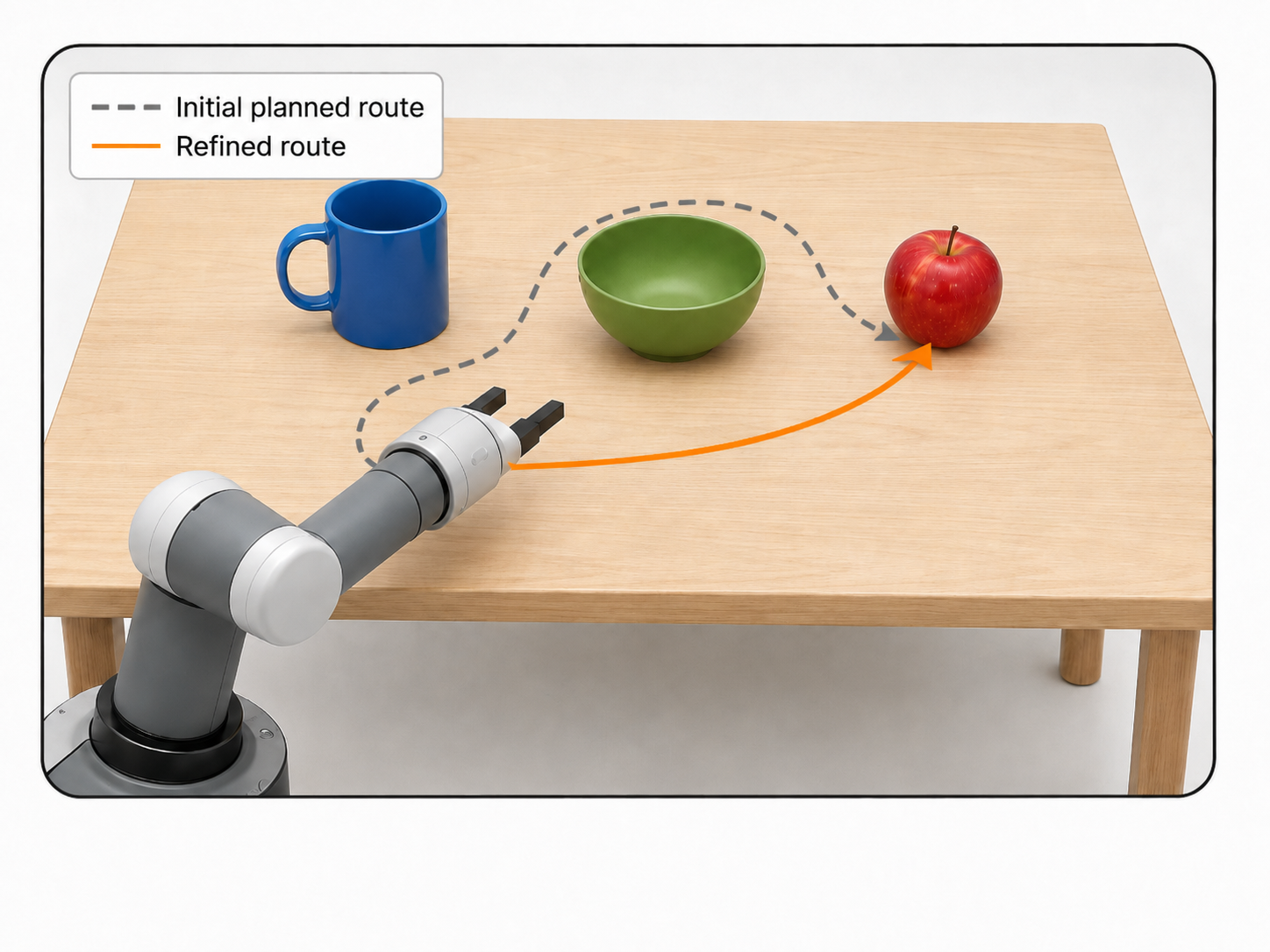}}
\par\small (3) Refine the Action
\end{minipage}
\caption{SG-LL grounds the current scene, anticipates a compact latent future
from policy-available context, and uses that lookahead to refine action
generation.}
\label{fig:teaser}
\end{figure}

We introduce \emph{Spatially Grounded Latent Lookahead} (SG-LL), summarized in
Figure~\ref{fig:teaser}. SG-LL first maps VLM Backbone features into Grounded Spatial State Tokens under supervision from a Frozen VGGT \cite{wang2025vggt} Spatial Teacher. It then
predicts compact, current-conditioned Future Latent Tokens from the grounded
state, instruction, robot state, horizon, and an Action Intent Representation
derived from initial Action Expert hidden states. Finally, zero-initialized
Spatial Grounding Residual Injection and Future-Latent Residual Fusion connect
these representations to the inherited policy. Together, these components implement a simple principle: ground the present, look ahead in latent space, and refine the action.
Because both residual pathways are zero-initialized, SG-LL preserves the pretrained policy computation at initialization while requiring no future observations, no online geometry model, and no additional 3D sensors at deployment.

We evaluate SG-LL at both the system and mechanism levels. SG-LL outperforms
the official $\pi_{0.5}$-LIBERO baseline across 10,030 paired LIBERO-Plus
episodes and an original-architecture $\pi_{0.5}$ baseline under matched
CALVIN D-to-D adaptation. Targeted interventions and trained controls examine
whether the policy behaviorally uses episode-matched future latents and
policy-derived action context, while distinguishing behavioral reliance from
architectural contribution. We further evaluate SG-LL on three real-robot
manipulation tasks.

Our contributions are as follows:
\begin{itemize}
  \item We introduce a spatial grounding interface that distills geometric priors from a frozen spatial teacher into policy-connected Grounded Spatial State Tokens, enabling control-relevant spatial grounding while retaining RGB-only deployment.

  \item We propose a decodable latent lookahead mechanism that predicts a compact, current-conditioned near-future representation from policy-available context and directly couples it to action generation through zero-initialized residual refinement.

 \item We validate SG-LL through paired simulation evaluations, targeted
mechanism interventions, and real-robot experiments, demonstrating improved
end-to-end control performance and showing that the trained policy makes
functional use of its predicted future latent.
\end{itemize}

\section{Related Work}

\subsection{Vision-Language-Action Models for Robot Control}

VLA policies combine pretrained vision-language representations with diverse
action-generation mechanisms, ranging from autoregressive action tokenization
to diffusion and flow matching for continuous control
\citep{brohan2022rt,zitkovich2023rt,kim2024openvla,team2024octo,
chi2025diffusion,liu2025rdt,black2024pi0,physicalintelligence2025pi05}.
Large-scale robot datasets and multimodal backbones further support transfer
across tasks and embodiments, while recent work improves action representations
and adaptation efficiency through latent actions, efficient tokenization, or
specialized policy heads
\citep{collaboration2023open,bousmalis2023robocat,li2024cogact,
bu2025univla,pertsch2025fast,kim2025fine}.

SG-LL targets a different axis: rather than redesigning the base VLA or its
action parameterization, it augments the inherited policy with control-relevant
spatial grounding and near-future predictive information that directly
participate in action generation.

\subsection{Spatial Representations and Geometric Grounding}

Spatial structure has been incorporated into robot policies through factorized
what/where pathways, volumetric representations, point-cloud inputs, and
explicit 3D feature encoders
\citep{shridhar2021cliport,pmlr-v205-shridhar23a,ze20243d}.
Recent VLA methods bring similar spatial inductive biases into pretrained
vision-language policies through 3D positional encoding, representation
alignment, or geometry-aware supervision
\citep{qu2025spatialvla,li2025spatial,wu2026pragmatic}.

SG-LL instead uses spatial grounding as part of a predictive control interface:
frozen geometric priors supervise a policy-connected current spatial state that
both feeds into the inherited action pathway and supports near-future latent
prediction. Spatial grounding is therefore coupled with predictive lookahead
and action refinement rather than introduced only as an additional spatial
representation.

\subsection{Predictive Representations and World Models}

Predictive representation learning captures temporal structure by predicting
future information in representation space rather than reconstructing every
image pixel \citep{assran2023self,bardes2024revisiting}. In robotics, future
observations, visual subgoals, latent actions, and predictive features have been
used for representation learning, action conditioning, and goal-directed
control
\citep{lynch2020learning,du2023learning,black2024zero,ye2025latent,zhao2025cot}.

More recent VLA-oriented methods align policy representations with future
visual features or latent futures, or generate future-oriented conditions and
imagined observations to guide action generation
\citep{zheng2025flare,su2026world,fang2026privileged,zhao2026frappe,
hu2024video,tian2025predictive,zhang2026dreamvla,sun2026vla,miao2026jepa,
li2026fomovla,xie2026forediffusion,yuan2026fast}.
Related world-model approaches learn action-conditioned dynamics for prediction,
rollout, and planning
\citep{hafner2019learning,hansen2022temporal,zhou2024dino,assran2025v},
including spatially structured models that connect geometric scene
representations to predictive robot control
\citep{zhen20243d,qiu2026making}.

SG-LL brings spatial grounding and predictive representation learning together
through a decodable, current-conditioned near-future latent. Rather than using
future prediction only as a pretraining signal or externally generated subgoal,
SG-LL directly integrates the predicted latent into Action Expert hidden states
through residual fusion, making near-future information actionable within the
pretrained VLA pathway.

\section{Methodology}

\subsection{Overview}

Let \(o_t\) denote the current multi-view RGB observation, \(\ell\) the
language instruction, \(s_t\) the robot state, and
\(A_t=a_{t:t+K-1}\) the action chunk. The base VLA encodes the current
multimodal context into a policy prefix, while the inherited flow-matching
Action Expert predicts a velocity field over the noisy action suffix that is
integrated to obtain \(A_t\). SG-LL augments this computation by forming
Grounded Spatial State Tokens \(S_t\), predicting a current-conditioned
near-future representation as Predicted Future Latent Tokens \(\hat Z_t\),
and using both representations to refine the inherited action-generation
pathway.

\begin{figure*}[t]
    \centering
    \includegraphics[width=\textwidth]{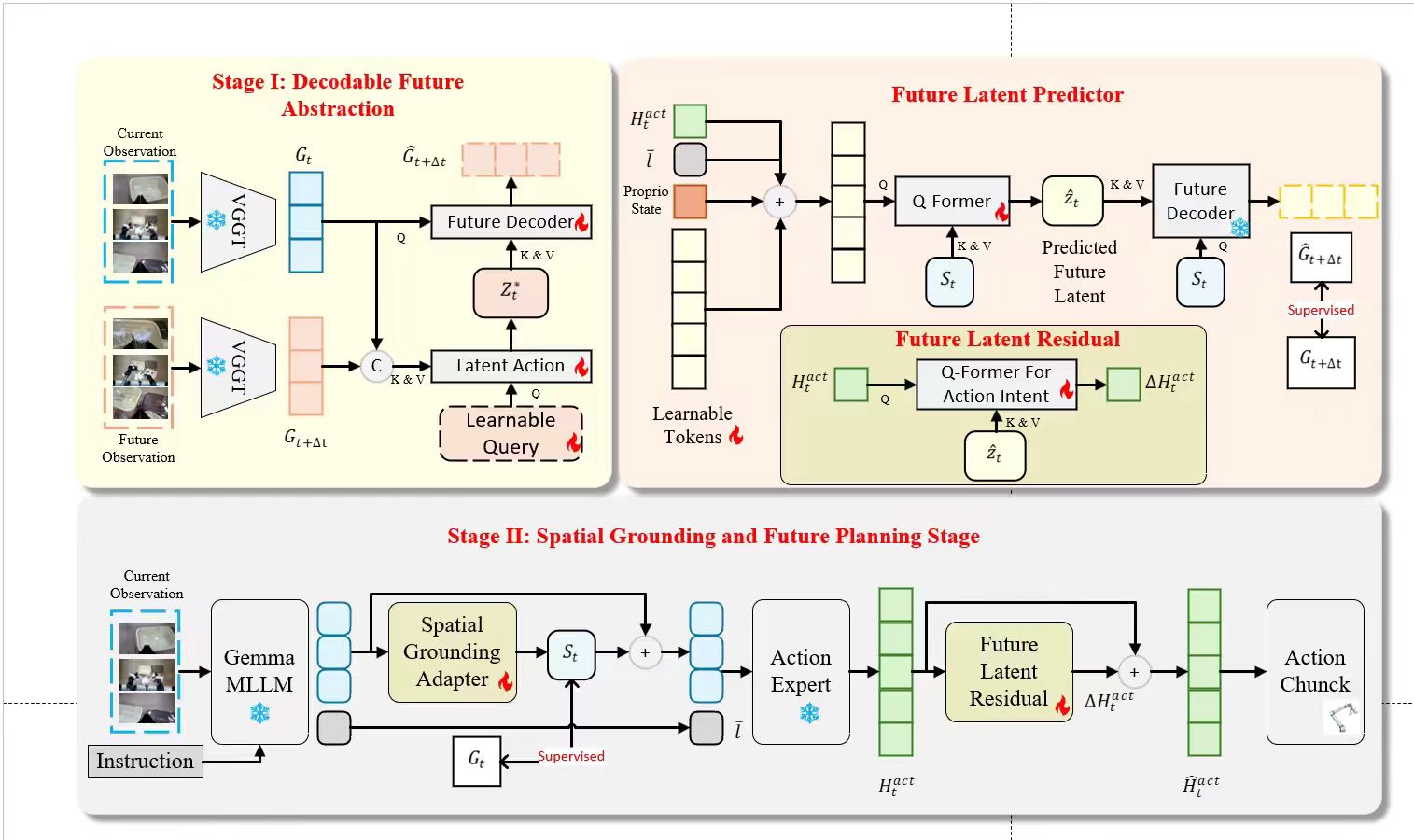}
    \caption{
    Overview of SG-LL. The Decodable Future Abstraction Stage learns a compact,
    current-conditioned future latent from paired Current and Future Spatial
    Target Tokens. The Spatial Grounding and Future Planning Stage grounds the
    current scene, predicts the near-future latent from policy-available
    context, and refines action generation through Spatial Grounding Residual
    Injection and Future-Latent Residual Fusion. Dashed regions denote
    training-only computation and are absent during deployment.
    }
    \label{fig:architecture}
\end{figure*}

Figure~\ref{fig:architecture} organizes SG-LL into two stages.
Stage I, the \emph{Decodable Future Abstraction Stage}, establishes a compact,
decodable future latent space from paired Current and Future Spatial Target
Tokens. Stage II, the \emph{Spatial Grounding and Future Planning Stage},
learns the policy-connected grounding, lookahead, and residual refinement
pathways. It forms \(S_t\), derives an Action Intent
Representation \(I_t^{\mathrm{act}}\) from initial Action Expert hidden states,
predicts \(\hat Z_t\) from the current policy context, and injects the grounded
present and predicted future into the inherited policy through residual
interfaces. In SG-LL, planning is realized as a single near-future latent
prediction that directly informs action generation.

\subsection{Decodable Future Abstraction Stage}

The first formal stage establishes a compact, decodable latent representation
of the near future. The Frozen VGGT Spatial Teacher \(\mathcal G\) maps the
current observation and its \(\Delta\)-step future counterpart to spatial
targets
\begin{equation}
G_t=\mathrm{sg}\!\left(\mathcal G(o_t)\right),\qquad
G_{t+\Delta}=\mathrm{sg}\!\left(\mathcal G(o_{t+\Delta})\right).
\label{eq:spatial-targets}
\end{equation}
where \(G_t,G_{t+\Delta}\in\mathbb R^{B\times N\times 2048}\) denote the
Current and Future Spatial Target Tokens, and
\(\mathrm{sg}(\cdot)\) denotes stop-gradient.

The Future Latent Encoder \(E_{\mathrm F}\) jointly encodes the pair into
compact Future Latent Tokens
\begin{equation}
Z_t^*=E_{\mathrm F}(G_t,G_{t+\Delta}),\qquad
Z_t^*\in\mathbb R^{B\times 8\times 256}.
\label{eq:future-latent-encoding}
\end{equation}
A Current-Conditioned Decoder \(D_{\mathrm{CC}}\) reconstructs the future
spatial target using the current spatial reference and the latent bottleneck:
\begin{equation}
\begin{aligned}
\widetilde G_{t+\Delta}
&=
D_{\mathrm{CC}}(G_t,Z_t^*),\\
\mathcal L_{\mathrm{abs}}
&=
d_{\mathrm{abs}}
(\widetilde G_{t+\Delta},G_{t+\Delta}).
\end{aligned}
\label{eq:stage1-reconstruction}
\end{equation}
Since \(D_{\mathrm{CC}}\) receives \(G_t\) directly, the bottleneck encourages
\(Z_t^*\) to capture future information that complements the current spatial
state. Accordingly, the latent encoding and current-conditioned reconstruction
in Eqs.~\eqref{eq:future-latent-encoding}--\eqref{eq:stage1-reconstruction}
establish \(Z_t^*\) as a compact representation of the near future relative
to the current spatial state.

After Stage I, \(E_{\mathrm F}\) and \(D_{\mathrm{CC}}\) are frozen and
serve as training-time supervision modules in Stage II. For each demonstrated
current--future pair, the frozen encoder constructs the target latent
\begin{equation}
Z_t^*
=
\mathrm{sg}\!\left(
E_{\mathrm F}(G_t,G_{t+\Delta})
\right).
\label{eq:target-future-latent}
\end{equation}
which is used by the condition-negative objective, while the frozen decoder
provides future-reconstruction supervision for the policy-side prediction.

\subsection{Spatial Grounding and Future Planning Stage}

\paragraph{Ground the present.}
The VLM Backbone produces current visual-language features, and the Spatial
Grounding Adapter maps its visual tokens into the spatial teacher space:
\begin{equation}
\begin{aligned}
H_t &= f_{\mathrm{VLM}}(o_t,\ell),\\
S_t &= A_{\mathrm{SG}}(H_t^{\mathrm{vis}}),
\qquad
S_t\in\mathbb R^{B\times N\times 2048}.
\end{aligned}
\label{eq:spatial-grounding}
\end{equation}
The grounded representation is aligned with the Current Spatial Target Tokens:
\begin{equation}
\mathcal L_{\mathrm{SG}}
=
d_{\mathrm{SG}}(S_t,G_t).
\label{eq:spatial-grounding-loss}
\end{equation}
Spatial Grounding Residual Injection further injects \(S_t\) into the policy
prefix through a learned residual pathway with a zero-initialized gate.
Thus, the grounded state is both geometrically supervised and directly
coupled to action generation, while the zero-initialized residual preserves
the pretrained policy function at initialization.

\paragraph{Form the Action Intent Representation.}
SG-LL evaluates the Action Expert once at the initial flow time
\(\tau=1\) using the initial noisy action suffix, obtaining Action Hidden
Tokens \(H_{t,\tau=1}^{\mathrm{act}}\). It then computes
\begin{equation}
I_t^{\mathrm{act}}
=
\mathrm{sg}\!\left(
\operatorname{MeanPool}
(H_{t,\tau=1}^{\mathrm{act}})
\right).
\label{eq:action-intent}
\end{equation}
This yields a compact policy-derived control context for conditioning the
latent lookahead.

\paragraph{Look ahead in latent space.}
Using the Action Intent Representation defined in
Eq.~\eqref{eq:action-intent}, the Future Latent Predictor conditions on the
Grounded Spatial State Tokens, pooled language representation \(\bar\ell\),
robot state \(s_t\), lookahead-horizon embedding \(e_\Delta\), and
policy-derived action context:
\begin{equation}
\hat Z_t
=
P_{\mathrm F}
(S_t,\bar\ell,s_t,e_\Delta,I_t^{\mathrm{act}}),
\qquad
\hat Z_t\in\mathbb R^{B\times 8\times 256}.
\label{eq:future-latent-prediction}
\end{equation}
The prediction in Eq.~\eqref{eq:future-latent-prediction} is computed from
policy-available context only. During Stage II training, the frozen
Current-Conditioned Decoder reconstructs the future spatial target from the
policy-side grounded state and predicted future latent:
\begin{equation}
\widetilde G_{t+\Delta}^{\mathrm{pred}}
=
D_{\mathrm{CC}}(S_t,\hat Z_t).
\label{eq:predicted-future-reconstruction}
\end{equation}
\begin{equation}
\mathcal L_{\mathrm F}
=
d_{\mathrm F}
\left(
\widetilde G_{t+\Delta}^{\mathrm{pred}},
G_{t+\Delta}
\right).
\label{eq:future-reconstruction-loss}
\end{equation}
Because \(S_t\) is aligned with the teacher-space current representation, the
frozen Stage-I decoder can use \(S_t\) as its current-state condition.
Together, Eqs.~\eqref{eq:predicted-future-reconstruction}--
\eqref{eq:future-reconstruction-loss} couple the grounded present \(S_t\)
and predicted future latent \(\hat Z_t\) to the demonstrated future spatial
target.

\paragraph{Refine the action.}
Future-Latent Residual Fusion cross-attends from the Action Hidden Tokens to
the Predicted Future Latent Tokens at each flow step:
\begin{equation}
H_{t,\tau}^{\mathrm{act}\prime}
=
H_{t,\tau}^{\mathrm{act}}
+
\alpha
F(H_{t,\tau}^{\mathrm{act}},\hat Z_t).
\label{eq:future-latent-fusion}
\end{equation}
where \(F\) denotes the fusion block and the learned gate \(\alpha\) is
initialized to zero. Through the residual refinement in
Eq.~\eqref{eq:future-latent-fusion}, the predicted latent directly modifies
the Action Expert hidden states at each flow step. The inherited Action Expert
output projection then predicts the flow velocity, which is integrated to
obtain \(A_t\). In contrast,
\(\widetilde G_{t+\Delta}^{\mathrm{pred}}\) is used only for training-time
future reconstruction.

\subsection{Learning Objectives and Deployment}

For Stage II training on LIBERO-Plus, we optimize
\begin{equation}
\mathcal L
=
\mathcal L_{\mathrm{flow}}
+0.06\mathcal L_{\mathrm{SG}}
+0.01\mathcal L_{\mathrm F}
+0.02\mathcal L_{\mathrm{neg}}.
\label{eq:stage2-objective}
\end{equation}
The overall objective in Eq.~\eqref{eq:stage2-objective} combines the inherited
flow-matching loss with spatial-grounding, future-reconstruction, and
condition-negative supervision. Here,
\(\mathcal L_{\mathrm{flow}}\) is the inherited flow-matching objective,
\(\mathcal L_{\mathrm{SG}}\) supervises current spatial grounding,
\(\mathcal L_{\mathrm F}\) supervises future reconstruction, and
\(\mathcal L_{\mathrm{neg}}\) is a margin-based condition-negative objective
that encourages the Future Latent Predictor to depend on language and
policy-derived action context.Specifically, it contrasts the matched future
latent against predictions obtained by cyclically shuffling
\(\bar\ell\) or \(I_t^{\mathrm{act}}\) within the minibatch.
Exact distance functions and the formulation of
\(\mathcal L_{\mathrm{neg}}\) are provided in the supplementary material.

In the primary LIBERO-Plus configuration, the VLM Backbone, Action Expert,
Frozen VGGT Spatial Teacher, Future Latent Encoder, and Current-Conditioned
Decoder are frozen, while the Spatial Grounding Adapter, Future Latent
Predictor, residual interfaces, and associated projections are optimized.

At inference, SG-LL forms \(S_t\), obtains \(I_t^{\mathrm{act}}\) from one
initial Action Expert suffix pass, predicts \(\hat Z_t\) once, and reuses the
result throughout flow denoising. Deployment retains the original input
interface \((o_t,\ell,s_t)\) and requires neither future observations nor the
VGGT teacher, Future Latent Encoder, or Current-Conditioned Decoder.

\section{Experiments}
\label{sec:experiments}

Our experiments address three questions.
(Q1) Does SG-LL improve end-to-end policy performance on LIBERO-Plus and
CALVIN under matched evaluation protocols?
(Q2) What does the learned future latent represent, does the trained policy
behaviorally use it, and how do the key design choices contribute?
(Q3) How does SG-LL perform on real-world manipulation tasks?

\begin{table*}[t]
\centering
\caption{
LIBERO-Plus robustness comparison (\%) across seven perturbation dimensions.
Prior methods are grouped by whether LIBERO-Plus data are used for policy adaptation. Published Overall scores follow the corresponding reported
benchmark results. The final two rows are our local evaluations at evaluation seed 7,
where Overall is the episode-weighted success rate over the complete
10,030-episode manifest rather than the arithmetic mean of the seven
perturbation columns. Boldface and underlining denote the best and second-best
results in each column, respectively.
}
\label{tab:main-results}
{\scriptsize
\setlength{\tabcolsep}{2.15pt}
\renewcommand{\arraystretch}{1.04}
\begin{tabular}{lcccccccc}
\toprule
Method & Camera & Robot & Language & Light & Background & Noise & Layout & Overall \\
\midrule

\multicolumn{9}{l}{\textit{Trained on LIBERO; zero-shot evaluation on LIBERO-Plus}} \\

UniVLA \citep{bu2025univla}
& 1.8 & 46.2 & 69.6 & 69.0 & 81.0 & 21.2 & 31.9 & 42.9 \\

OpenVLA-OFT \citep{kim2025fine}
& 56.4 & 31.9 & 79.5 & 88.7 & 93.3 & 75.8 & 74.2 & 69.6 \\

$\pi_0$ \citep{black2024pi0}
& 13.8 & 6.0 & 58.8 & 85.0 & 81.4 & 79.0 & 68.9 & 53.6 \\

$\pi_0$-Fast \citep{pertsch2025fast}
& 65.1 & 21.6 & 61.0 & 73.2 & 73.2 & 74.4 & 68.8 & 61.6 \\

WorldVLA \citep{cen2025worldvla}
& 0.1 & 27.9 & 41.6 & 43.7 & 17.1 & 10.9 & 38.0 & 25.0 \\

RIPT-VLA \citep{tan2025interactive}
& 55.2 & 31.2 & 77.6 & 88.4 & 91.6 & 73.5 & 74.2 & 68.4 \\

ResVLA \citep{zhong2026noise}
& 53.2 & 57.5 & 88.2 & 94.5 & 96.3 & 81.8 & 78.3 & 76.9 \\

VLA-JEPA \citep{sun2026vla}
& 63.3 & 67.1 & 85.4 & 95.6 & 93.6 & 66.3 & 85.1 & 79.5 \\

3DThinkVLA \citep{shi20263dthinkvla}
& 73.8 & 64.5 & 78.0 & \textbf{98.4} & 94.8 & 84.7 & 81.5 & 81.0 \\

TBD-VLA \citep{lee2026tbd}
& 87.8 & 60.4 & 77.4 & 95.8 & 88.8 & 89.9 & 84.4 & 83.5 \\

\midrule
\multicolumn{9}{l}{\textit{Adapted on LIBERO-Plus}} \\

OpenVLA-OFT+ \citep{fei2025libero}
& 92.8 & 30.3 & 85.8 & 94.9 & 93.9 & 89.3 & 77.6 & 79.6 \\

RoVLA \citep{luo2026rovla}
& \textbf{96.6} & 32.0 & \underline{91.5} & 95.9
& 96.1 & \underline{95.1} & 74.1 & 82.0 \\

ProGAL-VLA \citep{darabi2026progal}
& \underline{93.2} & 71.5 & \textbf{93.6} & 86.8
& 92.3 & 74.8 & \textbf{86.7} & 85.5 \\

ACoT-VLA \citep{zhong2026acot}
& \textbf{96.6} & 70.4 & 79.7 & 95.1
& \textbf{97.1} & \textbf{95.9} & 85.0 & \textbf{88.0} \\

\midrule
\multicolumn{9}{l}{\textit{Our local paired evaluation}} \\

$\pi_{0.5}$ \citep{physicalintelligence2025pi05}
& 72.6 & \underline{74.6} & 86.0 & \underline{96.8}
& \underline{96.4} & 87.5 & \underline{86.5} & 84.4 \\

\textbf{SG-LL (ours)}
& 80.5 & \textbf{77.5} & 85.5 & 95.1
& 95.4 & 91.2 & 85.4 & \underline{86.2} \\

\bottomrule
\end{tabular}}
\end{table*}

\subsection{Simulation Setup and Baselines}

\paragraph{LIBERO-Plus.}
We use LIBERO-Plus \citep{fei2025libero} as our primary simulation benchmark,
covering seven perturbation dimensions across the Spatial, Object, Goal, and
Long suites. Stage I is trained for 4,000 steps and then frozen. Stage II
adaptation initializes from the official $\pi_{0.5}$-LIBERO checkpoint,
freezes the VLM Backbone and Action Expert, fixes $\Delta=10$, and trains the
policy-side SG-LL modules for 10,000 steps with seed 42.

We use the official $\pi0.5$-LIBERO checkpoint as the primary reference because SG-LL freezes the pretrained VLM Backbone and Action Expert and adapts the policy exclusively through its newly introduced modules. Applying the same freezing rule to the original $\pi0.5$ architecture leaves no corresponding parameters to optimize, making the official checkpoint the natural frozen-policy reference. To separately control for exposure to LIBERO-Plus adaptation data, we additionally continue training the original $\pi0.5$ architecture for 10,000 steps using its standard trainable policy parameters. This continued-adaptation control reaches 88.38\% on Spatial, compared with 90.97\% for the official checkpoint and 93.88\% for SG-LL.

All local evaluations use the same preprocessing, normalization, inference
settings, and episode manifests. Results are paired by episode and evaluated
with exact McNemar tests and paired-bootstrap confidence intervals. Overall
scores are episode-weighted over the complete 10,030-episode manifest.

\paragraph{CALVIN.}
We use CALVIN \citep{mees2022calvin} as a secondary cross-benchmark evaluation under
a matched D-to-D adaptation protocol. SG-LL and the original-architecture
$\pi_{0.5}$ baseline initialize from the same $\pi_{0.5}$-base checkpoint
and use identical Task-D data and 10,000-step adaptation budgets. Both freeze
the VLM Backbone and train the Action Expert. Evaluation uses the same fixed
1,000 five-task sequences with a 360-step episode limit and replanning
interval 5.

\begin{figure*}[t]
\centering

\begin{minipage}[t]{0.49\textwidth}
\centering
\includegraphics[width=0.28\linewidth]{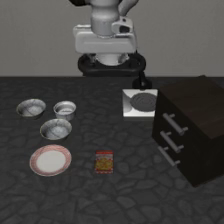}\hfill
\includegraphics[width=0.28\linewidth]{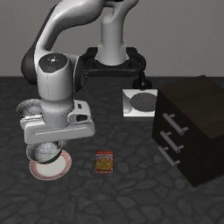}\hfill
\includegraphics[width=0.28\linewidth]{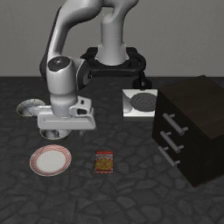}
\vspace{0.7mm}

{\small (a) Experimental setups on LIBERO-Plus}
\end{minipage}
\hfill
\begin{minipage}[t]{0.49\textwidth}
\centering
\includegraphics[width=0.28\linewidth]{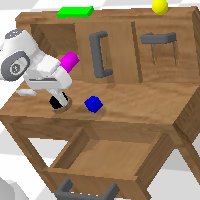}\hfill
\includegraphics[width=0.28\linewidth]{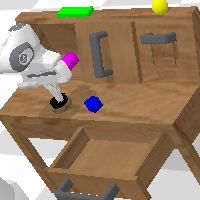}\hfill
\includegraphics[width=0.28\linewidth]{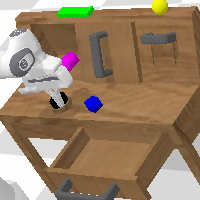}
\vspace{0.7mm}

{\small (b) Experimental setups on CALVIN}
\end{minipage}

\vspace{1.5mm}

\begin{minipage}[t]{0.32\textwidth}
\centering
\includegraphics[width=\linewidth]{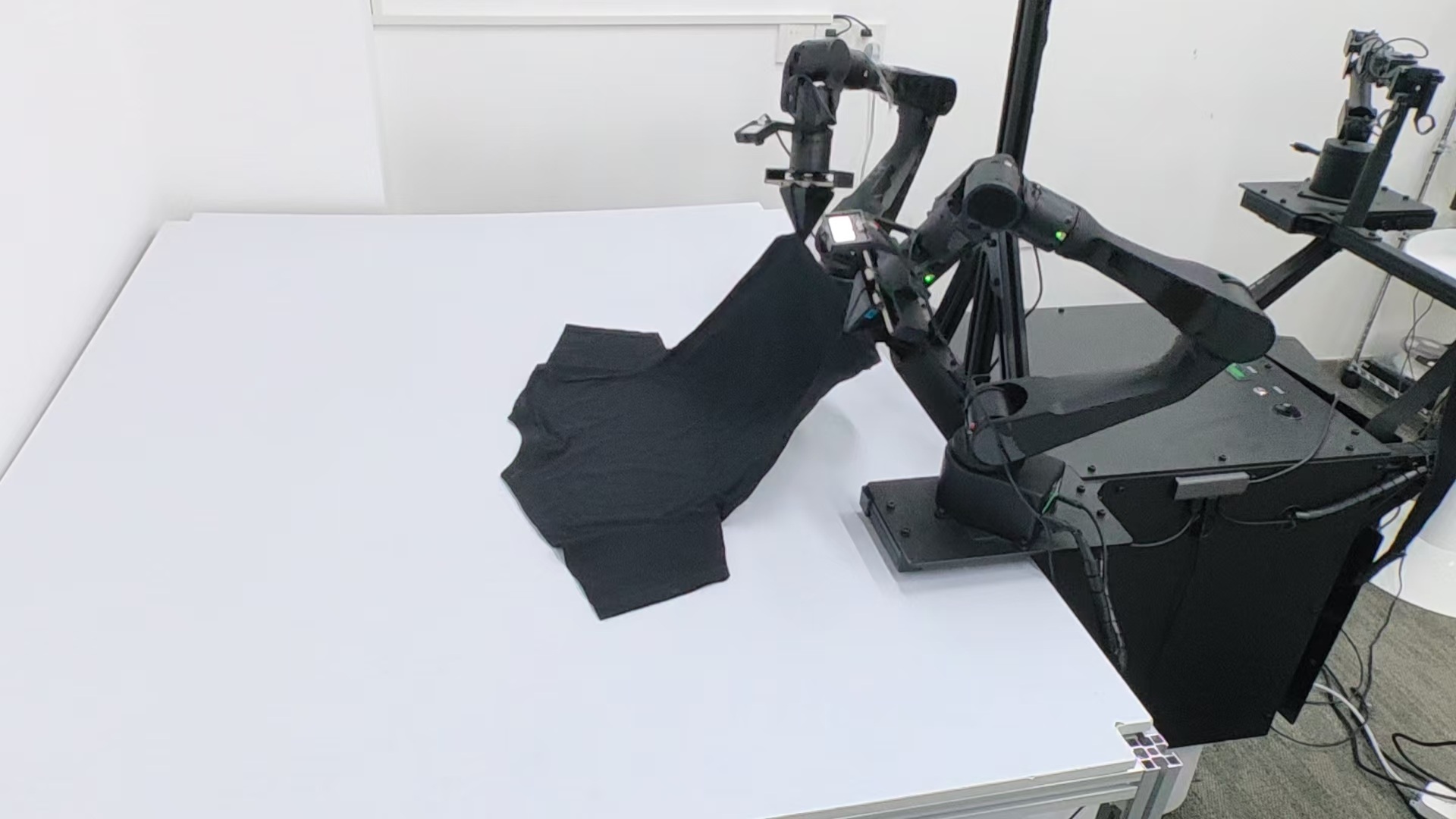}
\vspace{0.4mm}

{\small Fold clothes}
\end{minipage}
\hfill
\begin{minipage}[t]{0.32\textwidth}
\centering
\includegraphics[width=\linewidth]{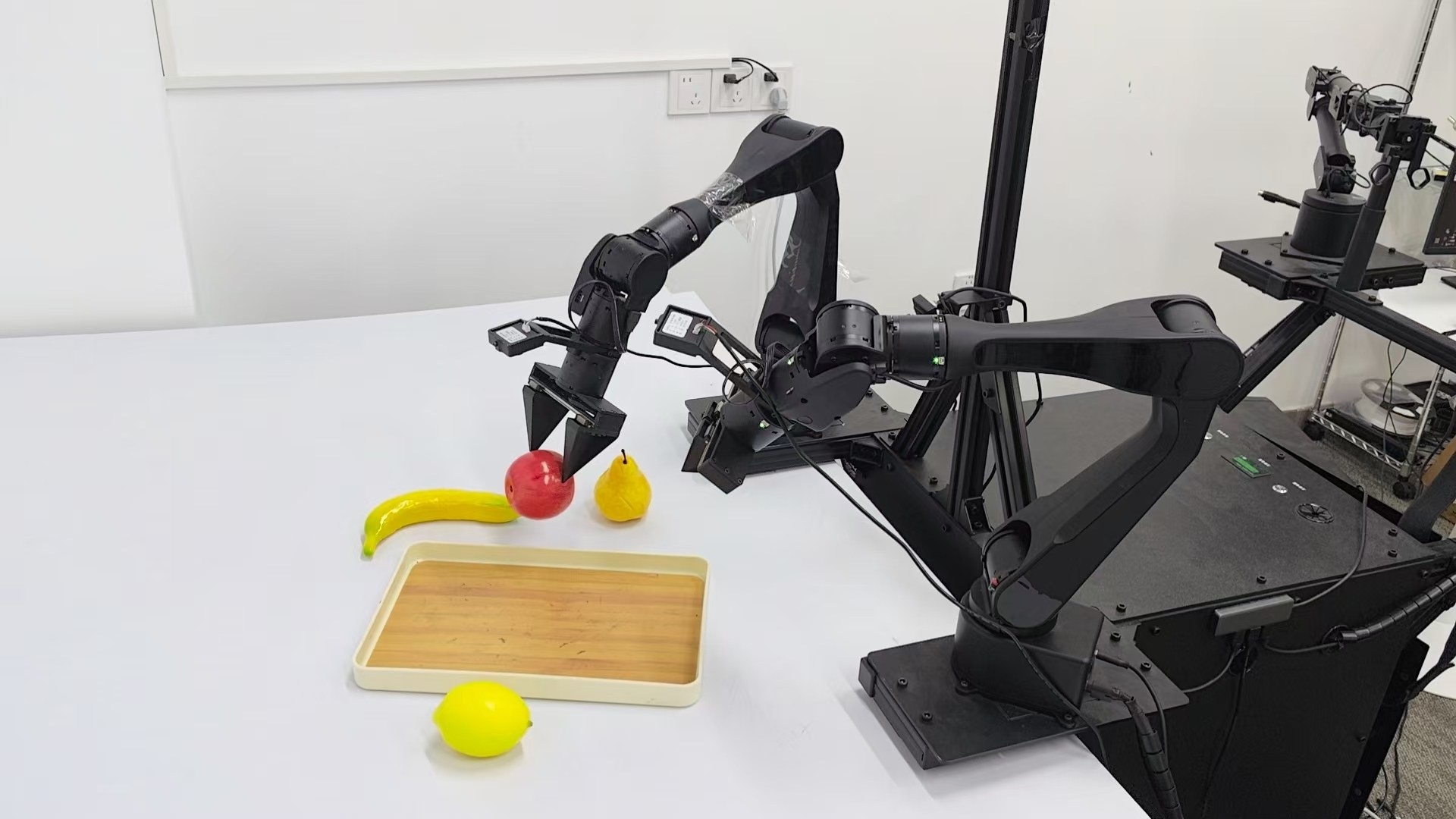}
\vspace{0.4mm}

{\small Pick up fruit}
\end{minipage}
\hfill
\begin{minipage}[t]{0.32\textwidth}
\centering
\includegraphics[width=\linewidth]{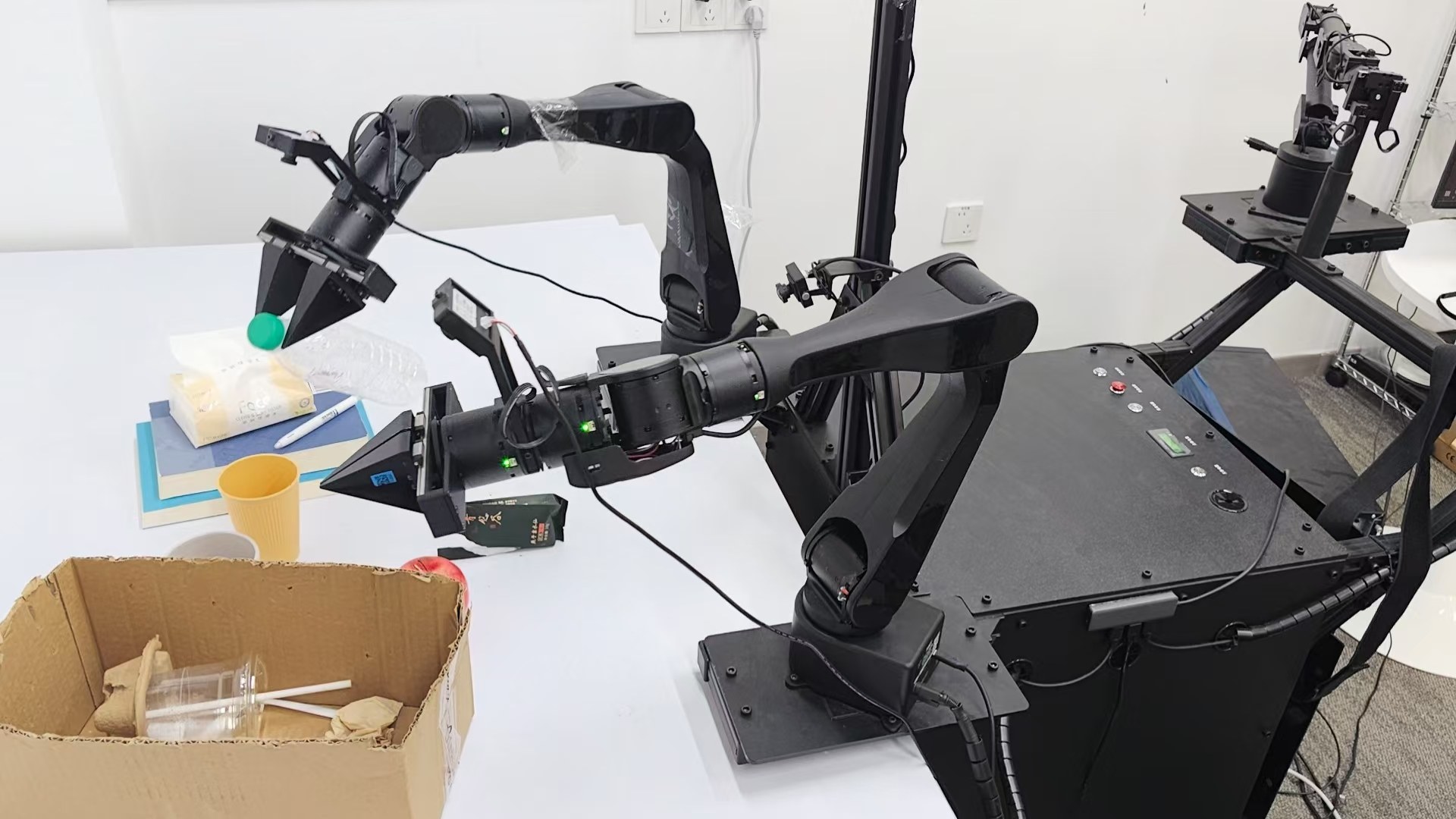}
\vspace{0.4mm}

{\small Organize desk}
\end{minipage}

\vspace{0.7mm}

{\small (c) Experimental setups on the AIRBOT TOK2 platform}

\caption{
Experimental setups on LIBERO-Plus, CALVIN, and the AIRBOT TOK2 platform.
We evaluate SG-LL on two simulation benchmarks and one real-world environment.
The real-world system consists of two six-DoF AIRBOT Play arms with
two-finger grippers, two 720p wrist-mounted RGB cameras, and one 1080p
overhead RGB camera.
}
\label{fig:experimental-setups}
\end{figure*}

\subsection{Simulation Evaluation}

\begin{figure}[!t]
\centering
\includegraphics[width=0.85\linewidth]{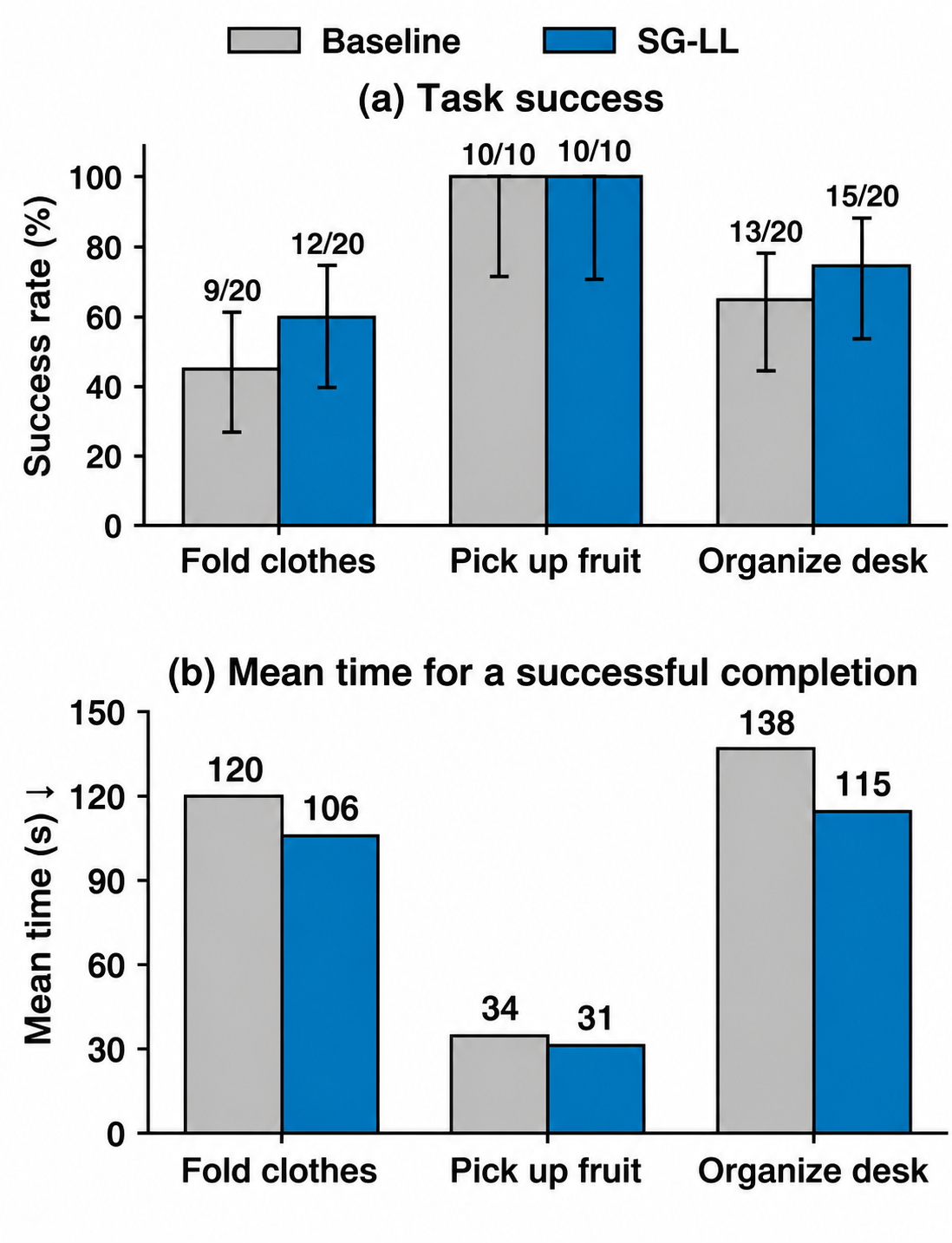}
\caption{
Real-world task performance over 20, 10, and 20 trials for fold clothes,
pick up fruit, and organize desk, respectively. Error bars denote Wilson
95\% binomial intervals; completion time is averaged over successful trials.
}
\label{fig:real-world-results}
\end{figure}

\paragraph{LIBERO-Plus robustness.}
Table~\ref{tab:main-results} compares SG-LL with recent VLA policies across
the seven LIBERO-Plus perturbation dimensions, distinguishing zero-shot
evaluation after LIBERO training from methods adapted on LIBERO-Plus.
Among adapted methods, SG-LL achieves 86.21\% overall success and the
highest Robot score (77.5\%), while remaining competitive with ACoT-VLA
(88.0\%).

On the identical 10,030-episode paired manifest, SG-LL improves the official
$\pi_{0.5}$-LIBERO baseline from 84.45\% to 86.21\%, a $+1.765$-pp gain
(exact McNemar $p=1.88\times10^{-7}$; paired-bootstrap 95\% CI
$[+1.147,+2.423]$ pp). The gain is consistent across training seeds:
Spatial success reaches 93.88\%, 92.80\%, and 92.13\% for seeds 42--44,
respectively, versus 90.97\% for the official baseline. Seed-specific paired
statistics are provided in the supplementary material.

\paragraph{CALVIN D-to-D adaptation.}
Under the matched protocol above, SG-LL increases the mean number of completed
subtasks per sequence from 0.580 to 0.725, a 25.0\% relative improvement
(paired-bootstrap 95\% CI $[+0.075,+0.217]$).
Detailed chain-length results are provided in the supplementary material.

\subsection{Mechanism Analysis and Ablations}
\label{sec:analysis}
\label{sec:ablations}

\paragraph{What does the learned future latent represent?}
We first examine the representation learned by the Decodable Future
Abstraction Stage. A frozen-stage probe predicts $Z^*$ from future spatial
features with a loss of 0.011462, compared with 0.297392 when using only
current features, showing that the latent is strongly future-informative.
At the same time, future decoding depends on both the latent and the current
reference: joint, current-only, and token-only decoding give losses of
0.074566, 0.093445, and 0.124070, respectively. Replacing the correct current
reference with zeros further increases the frozen-decoder loss from 0.076249
to 0.789774. Together, these results characterize $Z^*$ as a
\emph{future-dominant, current-complementary} representation: it carries
substantial information about the near future while being designed to decode
that future relative to the current spatial state. A sweep over 1--16 latent tokens shows the same future-dominant,
current-complementary pattern, suggesting that this behavior is not simply
determined by bottleneck size.

\paragraph{Does the trained policy use the predicted lookahead?}
We intervene on the same seed-42 SG-LL checkpoint over the fixed
2,402-episode Spatial manifest. As shown in
Table~\ref{tab:inference-interventions}, disabling residual fusion or replacing
the predicted latent with random alternatives substantially degrades
performance. Norm-matched random tokens behave similarly to unconstrained
random tokens, whereas cross-episode tokens retain partial performance,
showing that SG-LL relies on episode-specific latent content rather than merely
residual magnitude. Zeroing $I_t^{\mathrm{act}}$ also significantly reduces
success, confirming that lookahead prediction uses policy-derived action
context.

\begin{table}[t]
\centering
\caption{
Fixed-policy interventions on the 2,402-episode LIBERO-Plus Spatial manifest
using the same SG-LL checkpoint. $\Delta$ is relative to normal inference;
$p$ denotes the exact McNemar test.
}
\label{tab:inference-interventions}
{\small
\setlength{\tabcolsep}{3.5pt}
\begin{tabular}{lrrr}
\toprule
Setting & Success (\%) & $\Delta$ (pp) & $p$ \\
\midrule
Normal SG-LL              & \textbf{93.88} & --    & -- \\
$\alpha=0$                & 90.55 & -3.33 & $6.00{\times}10^{-9}$ \\
Random latent             & 90.34 & -3.54 & $2.65{\times}10^{-9}$ \\
Norm-matched random       & 90.22 & -3.66 & $2.78{\times}10^{-9}$ \\
Cross-episode latent      & 92.34 & -1.54 & $4.16{\times}10^{-3}$ \\
Zero $I_t^{\mathrm{act}}$ & 92.13 & -1.75 & $1.21{\times}10^{-3}$ \\
\bottomrule
\end{tabular}}
\end{table}

\paragraph{What is the architectural contribution of residual fusion?}
We compare SG-LL with a separately trained control that disables
Future-Latent Residual Fusion throughout Stage II training and evaluation,
while retaining the same Spatial Grounding Adapter, data protocol, and
training budget. As shown in Table~\ref{tab:fusion-control}, the contribution
of fusion is task-dependent: it significantly improves Spatial, is neutral on
Object, and degrades Goal and Long, resulting in no aggregate advantage over
the fusion-disabled control.

Together with Table~\ref{tab:inference-interventions}, this reveals an
important distinction: a trained SG-LL policy behaviorally relies on the
lookahead pathway, whereas a model trained without that pathway can partially
compensate through its remaining components. Thus, fixed-policy intervention
and independently trained controls measure different aspects of the fusion
mechanism.

\begin{table}[t]
\centering
\caption{
Comparison with the trained Future-Latent Residual Fusion-disabled control.
Both policies use the same data protocol and Stage II training budget.
Aggregate denotes the episode-weighted result over all 10,030 episodes.
$\Delta$ denotes SG-LL minus the fusion-disabled control.
}
\label{tab:fusion-control}
{\small
\setlength{\tabcolsep}{3.2pt}
\begin{tabular}{lrrrr}
\toprule
Suite & Fusion off & SG-LL & $\Delta$ (pp) & $p$ \\
\midrule
Spatial   & 92.21 & \textbf{93.88} & +1.67 & $1.70{\times}10^{-3}$ \\
Object    & \textbf{90.39} & 90.31 & -0.08 & $9.38{\times}10^{-1}$ \\
Goal      & \textbf{81.63} & 80.93 & -0.70 & $2.68{\times}10^{-1}$ \\
Long      & \textbf{83.17} & 80.23 & -2.94 & $4.45{\times}10^{-4}$ \\
\midrule
Aggregate & \textbf{86.75} & 86.21 & -0.54 & $9.24{\times}10^{-2}$ \\
\bottomrule
\end{tabular}}
\end{table}

\paragraph{Does current-conditioned future abstraction matter?}
Finally, we compare SG-LL with a future-frame latent control that removes the
sample-specific Current Spatial Target Tokens from Stage I while retaining
the same latent size, Future Latent Predictor, residual fusion, data, and
10,000-step Stage II training budget. The future-frame variant reaches
$8588/10030=85.62\%$, improving over the official $\pi_{0.5}$-LIBERO policy
by $1.176$ pp ($p=0.0003707$). SG-LL further reaches
$8647/10030=86.21\%$, an additional $0.588$-pp improvement over the
future-frame variant (paired 95\% CI $[-0.010,+1.186]$ pp,
$p=0.061023$).

This comparison shows that making compact future information actionable is
already a strong design choice. Conditioning the future abstraction on the
current spatial state yields the best numerical performance, although the
additional gain over direct future-frame latent prediction is modest.
Together with the frozen-stage probes, the result supports the role of
current-conditioned lookahead without requiring the latent to encode a pure
state difference.

\paragraph{Scope of the lookahead.}
The main model uses a fixed short-horizon offset of $\Delta=10$. The strong
Spatial gain and weaker Long result suggest that the current short-horizon
lookahead may be particularly effective for local manipulation decisions,
while longer multi-stage tasks may benefit from a broader temporal scope.
We leave multi-horizon prediction and adaptive lookahead as directions for
future study.

\FloatBarrier

\subsection{Real-World Experiments}
\label{sec:real-world}

We compare SG-LL with task-matched $\pi0.5$ baselines on three
real-robot tasks: \emph{fold clothes}, \emph{pick up fruit}, and \emph{organize desk}.
The baseline is obtained by fine-tuning the official $\pi_{0.5}$ model using
the official training recipe, with the same task demonstrations and training
budget as SG-LL. The datasets contain 6.5, 0.26, and 5.65 hours of
demonstrations, respectively. Experiments are conducted on the AIRBOT TOK2
platform shown in Figure~\ref{fig:experimental-setups}; detailed training
settings are provided in the supplementary material.

As shown in Figure~\ref{fig:real-world-results}, SG-LL improves success on
fold clothes and organize desk while matching perfect fruit-picking success,
and also achieves lower mean completion time among successful trials across
all three tasks. Given the modest number of trials, we treat these results as task-level real-world
evidence rather than aggregate statistical comparisons.

Across all three real-world tasks, training and evaluation use independently
sampled scene configurations from the same task-specific data distribution.
Object arrangements, spatial placements, and other initial conditions therefore
differ between training and evaluation, while the object sets and task
definitions remain fixed. These experiments evaluate in-distribution
generalization to unseen scene configurations rather than out-of-distribution
or novel-object generalization.

\section{Conclusion}

We introduced SG-LL, which augments VLA policies with spatially grounded,
current-conditioned latent lookahead that directly refines action generation.
Geometric teacher targets and future observations are used only during
training, preserving the standard RGB-based deployment interface.
Experiments on LIBERO-Plus, CALVIN, and real-world manipulation demonstrate
system-level improvements, while targeted interventions show that trained
SG-LL policies behaviorally use the predicted lookahead.

\bibliography{aaai2027}

\clearpage
\appendix
\begin{center}
{\LARGE\bf Supplementary Material\par}
\end{center}

\section{Coverage of Main-Paper Supplement References}

The main paper explicitly defers four groups of details to this supplement.
Table~\ref{tab:coverage} provides a direct index so that each item can be
located without searching the full document.

\begin{table*}[t]
\centering
\caption{Coverage of statements explicitly deferred from the main paper.}
\label{tab:coverage}
\small
\begin{tabular}{p{0.29\textwidth}p{0.60\textwidth}}
\toprule
Main-paper reference & Location in this supplement \\
\midrule
Exact distance functions and $\mathcal L_{\mathrm{neg}}$ &
The Full Method Details section gives the masked spatial distances, latent
distance, shuffled-condition construction, margin, and full Stage-II objective. \\
Seed-specific paired statistics &
The Training-Seed Checks subsection reports Spatial results for seeds 42--44
and the additional Object, Goal, and LIBERO-10 checks. \\
Detailed CALVIN chain-length results &
The CALVIN D-to-D protocol and CALVIN Full-1,000 Results subsections report
completion rates for chain lengths one through five with paired exact tests. \\
Detailed real-robot training settings &
The Real Robot protocol and Real-Robot Results subsections report the platform,
cameras, demonstration hours, batch sizes, optimization steps, trial counts,
success-rate intervals, and the scope of the archived task-level records. \\
\bottomrule
\end{tabular}
\end{table*}

\section{Scope and Terminology}

This supplement documents Spatially Grounded Latent Lookahead (SG-LL), its
implementation, and the completed experiments supporting the submission.
Frozen-feature probes motivate the term \emph{current-conditioned future
latent}: the target is strongly future-informative, while the current reference
and latent remain complementary for future decoding.  SG-LL implements a
single fixed-horizon latent lookahead that directly refines the inherited
action pathway.

All quantitative entries below are drawn from completed, archived artifacts.
Historical implementation identifiers such as \texttt{Route A},
\texttt{Scheme4}, and \texttt{delta} refer to SG-LL in the released code
(\url{https://github.com/Cqhddyb/SG-LL}).

\section{Full Method Details}

\subsection{Notation and Exact Distances}

Let $o_t$, $\ell$, $s_t$, and $A_t$ denote the current multi-view
observation, language instruction, normalized robot state, and action chunk.
The selected horizon and demonstrated-future offset are both $h=10$.  A frozen
spatial teacher maps current and future observations to
$G_t,G_{t+h}\in\mathbb{R}^{B\times N\times2048}$, where $N=768$ before
masking and 512 tokens are valid in the recorded three-view LIBERO-Plus
cache.  All teacher targets are detached.

For a token mask $M\in\{0,1\}^{B\times N}$, define the masked mean
$\langle q\rangle_M=\sum_{bn}M_{bn}q_{bn}/\max(1,\sum_{bn}M_{bn})$.  The
implemented absolute, cosine, and pairwise-relation distances are
\begin{align}
d_{\mathrm{abs}}(X,Y;M)
&=\left\langle \frac{1}{D}\sum_d
\operatorname{SmoothL1}(X_{bnd},Y_{bnd})\right\rangle_M,\\
d_{\cos}(X,Y;M)
&=\left\langle1-\frac{X_{bn}^{\top}Y_{bn}}
{\lVert X_{bn}\rVert_2\lVert Y_{bn}\rVert_2}\right\rangle_M,\\
d_{\mathrm{rel}}(X,Y;M)
&=\frac{\sum_{bnm}M_{bn}M_{bm}
\left|\bar X_{bn}^{\top}\bar X_{bm}-
\bar Y_{bn}^{\top}\bar Y_{bm}\right|}
{\max(1,\sum_{bnm}M_{bn}M_{bm})},
\end{align}
where $\bar X$ and $\bar Y$ are normalized along the feature dimension.  The
relation term uniformly subsamples at most 128 token positions when necessary.
The current spatial-grounding and future-reconstruction distances are
\begin{align}
d_{\mathrm{SG}}(X,Y;M)&=d_{\cos}+0.1d_{\mathrm{abs}}+0.05d_{\mathrm{rel}},\\
d_{\mathrm F}(X,Y;M)&=d_{\mathrm{abs}}+0.1d_{\cos}+0.05d_{\mathrm{rel}}.
\end{align}
Stage I used the same implementation with relation weight set explicitly to
zero, i.e., $d_{\mathrm F}^{\mathrm{I}}=d_{\mathrm{abs}}+0.1d_{\cos}$.
For latent tokens $X,Y\in\mathbb{R}^{B\times8\times256}$,
\begin{equation}
\begin{aligned}
d_Z(X,Y)={}&\mean\operatorname{SmoothL1}(X,Y)\\
&+0.1\mean_{bm}\left[1-\cos(X_{bm},Y_{bm})\right].
\end{aligned}
\end{equation}
Here $\sg(\cdot)$ means stop-gradient: its value is preserved and its
derivative is identically zero.

\subsection{Stage I: Decodable Future Abstraction}

The frozen teacher first produces $\sg(G_t)$ and $\sg(G_{t+h})$.  The Future
Latent Encoder constructs
\begin{equation}
Z^*=E_{\mathrm F}(\sg(G_t),\sg(G_{t+h})),\qquad
Z^*\in\mathbb{R}^{B\times8\times256}.
\end{equation}
The Current-Conditioned Decoder predicts
\begin{equation}
\widetilde G_{t+h}=D_{\mathrm{CC}}(\sg(G_t),Z^*),
\end{equation}
and is trained with
\begin{equation}
\mathcal L_{\mathrm I}=d_{\mathrm F}^{\mathrm I}
(\widetilde G_{t+h},\sg(G_{t+h});M).
\end{equation}
The selected run sets the optional identity loss to zero.  Therefore the
decoder's zero-initialized final residual projection provides initialization
identity, but no separate identity objective contributes to the selected
Stage-I checkpoint.

The encoder projects 2048-dimensional current/future tokens to hidden width
256, adds learned current/future type embeddings, position and horizon
embeddings, concatenates the two streams, and applies two layers of
learned-query self-attention followed by cross-attention.  Eight learned
queries are projected to width 256.  The decoder projects both current tokens
and latent tokens to width 256 and applies two pre-LayerNorm, 8-head
cross-attention blocks.  A LayerNorm--linear--GELU--linear residual head maps
back to 2048 dimensions; its final linear layer is initialized to exact zero.
All dropout probabilities are zero in the selected checkpoint.

\begin{algorithm}[t]
\caption{Stage I: decodable future abstraction}
\begin{algorithmic}[1]
\Require Paired observations $(o_t,o_{t+h})$, fixed horizon $h=10$
\State $G_t\gets\sg(T_{\mathrm{spatial}}(o_t))$
\State $G_{t+h}\gets\sg(T_{\mathrm{spatial}}(o_{t+h}))$
\State $Z^*\gets E_{\mathrm F}(G_t,G_{t+h},h)$
\State $\widetilde G_{t+h}\gets D_{\mathrm{CC}}(G_t,Z^*,h)$
\State update $E_{\mathrm F},D_{\mathrm{CC}}$ using
$d_{\mathrm F}^{\mathrm I}(\widetilde G_{t+h},G_{t+h};M)$
\end{algorithmic}
\end{algorithm}

\subsection{Stage II: Ground, Look Ahead, and Refine}

The frozen VLM produces visual/language prefix hidden states.  The Spatial
Grounding Adapter maps each visual token through
LayerNorm--linear--GELU--linear to
$S_t\in\mathbb{R}^{B\times N\times2048}$.  It is supervised by
\begin{equation}
\mathcal L_{\mathrm{SG}}=d_{\mathrm{SG}}(S_t,\sg(G_t);M).
\end{equation}
The same $S_t$ is projected through LayerNorm--linear--GELU--linear to the VLM
width and added to the original visual prefix through a channel-wise
$\tanh$ gate initialized to an exact zero vector.  This is Spatial Grounding
Residual Injection; with its gate at zero, the prefix and original policy path
are unchanged.

To condition lookahead on the policy's current control hypothesis, the action
expert processes the initial $\tau=1$ noisy suffix once and produces
$H_{t,\tau=1}^{\mathrm{act}}\in
\mathbb{R}^{B\times H_A\times D_{\mathrm{AE}}}$.  The Action Intent
Representation is
\begin{equation}
I_t^{\mathrm{act}}=\sg\left(
\frac{1}{H_A}\sum_{j=1}^{H_A}H_{t,\tau=1,j}^{\mathrm{act}}
\right)\in\mathbb{R}^{B\times D_{\mathrm{AE}}}.
\end{equation}
It provides a compact, policy-derived control context for the future-latent predictor.

The Future Latent Predictor projects $S_t$, masked-mean pooled language,
robot state, $I_t^{\mathrm{act}}$, and the fixed horizon to hidden width 1024.
Their projected condition sum is added to eight learned queries.  Three
pre-LayerNorm 8-head cross-attention/4$\times$-MLP blocks read $S_t$, followed
by LayerNorm--linear--GELU--linear projection to
$\hat Z_t\in\mathbb{R}^{B\times8\times256}$.  The frozen decoder gives
\begin{equation}
\begin{aligned}
\widehat G_{t+h}&=D_{\mathrm{CC}}(S_t,\hat Z_t),\\
\mathcal L_{\mathrm F}&=d_{\mathrm F}
(\widehat G_{t+h},\sg(G_{t+h});M).
\end{aligned}
\end{equation}
The frozen encoder is still used in Stage II to form the detached target
$Z^*=E_{\mathrm F}(\sg(G_t),\sg(G_{t+h}))$ needed by the condition-negative
loss.  Future observations remain target-only.

For conditions $\mathcal C=\{\bar\ell,I_t^{\mathrm{act}}\}$, negative
conditions are constructed deterministically by cyclically shifting one
condition by one row along the minibatch dimension while all other inputs
remain matched.  Let $\hat Z_t^{(-c)}$ be the resulting predictor output.  The
exact loss is
\begin{equation}
\begin{aligned}
\mathcal L_{\mathrm{neg}}
=\frac{1}{|\mathcal C|}\sum_{c\in\mathcal C}\big[&0.05
+d_Z(\hat Z_t,\sg(Z^*))\\
&-d_Z(\hat Z_t^{(-c)},\sg(Z^*))\big]_+.
\end{aligned}
\end{equation}
Language and action-condition hinges are averaged.  For batch size one the
shift is invalid and this term is exactly zero.  The selected objective is
\begin{equation}
\mathcal L=\mathcal L_{\mathrm{flow}}+0.06\mathcal L_{\mathrm{SG}}
+0.01\mathcal L_{\mathrm F}+0.02\mathcal L_{\mathrm{neg}},
\end{equation}
with no standalone direct $d_Z(\hat Z_t,Z^*)$ regression term.

Future-Latent Residual Fusion linearly maps the predicted latent to
$D_{\mathrm{AE}}$, performs pre-LayerNorm 8-head cross-attention from current
action hidden tokens to this memory, and applies a
LayerNorm--linear($2D_{\mathrm{AE}}$)--GELU--linear update.  A channel-wise
$\tanh$ gate initialized exactly to zero scales the update:
\begin{equation}
H_t^{\mathrm{act}\prime}=H_t^{\mathrm{act}}+
\alpha\,\tanh(g_{\mathrm F})\odot
U(\operatorname{XAttn}(H_t^{\mathrm{act}},\hat Z_t)),
\end{equation}
where $\alpha=1$ normally and $g_{\mathrm F}=0$ at initialization.  The
original action output projection maps $H_t^{\mathrm{act}\prime}$ to flow
velocity.  No implicit tensor broadcasting is used at module boundaries.

\begin{algorithm}[t]
\caption{Stage II: policy adaptation}
\begin{algorithmic}[1]
\Require Current input $(o_t,\ell,s_t)$, target-only $o_{t+h}$, actions $A_t$
\State obtain frozen prefix hidden and compute $S_t$
\State compute $G_t,G_{t+h}$ with the frozen spatial teacher and detach
\State $Z^*\gets\sg(E_{\mathrm F}(G_t,G_{t+h}))$
\State compute $\mathcal L_{\mathrm{SG}}$ from $(S_t,G_t)$
\State inject zero-gated spatial residual into the prefix
\State run the initial suffix; $I_t^{\mathrm{act}}\gets\sg(\mean H_{t,1}^{\mathrm{act}})$
\State $\hat Z_t\gets P_{\mathrm F}(S_t,\bar\ell,s_t,h,I_t^{\mathrm{act}})$
\State decode $\widehat G_{t+h}\gets D_{\mathrm{CC}}(S_t,\hat Z_t)$
\State construct shifted language/action negatives and $\mathcal L_{\mathrm{neg}}$
\State fuse $\hat Z_t$ into action hidden and compute $\mathcal L_{\mathrm{flow}}$
\State update only SG-LL adapters using the complete objective
\end{algorithmic}
\end{algorithm}

\subsection{Inference}

Inference receives only $(o_t,\ell,s_t)$.  It builds the spatially grounded
prefix key--value cache once, obtains $I_t^{\mathrm{act}}$ from one initial
$\tau=1$ suffix pass, predicts $\hat Z_t$ once outside the flow-denoising
loop, and reuses both the cache and latent for all ten flow steps.  The Frozen
VGGT Spatial Teacher, future observation, Future Latent Encoder, and
Current-Conditioned Decoder are absent online.

\begin{algorithm}[t]
\caption{Deployment-time action generation}
\begin{algorithmic}[1]
\Require Current RGB $o_t$, instruction $\ell$, state $s_t$
\State build grounded prefix KV cache and retain $(S_t,\bar\ell)$
\State initialize action noise and evaluate one $\tau=1$ suffix
\State $I_t^{\mathrm{act}}\gets\sg(\mean H_{t,1}^{\mathrm{act}})$
\State $\hat Z_t\gets P_{\mathrm F}(S_t,\bar\ell,s_t,h,I_t^{\mathrm{act}})$
\For{each inherited flow-denoising step}
  \State evaluate action suffix using the cached prefix
  \State apply Future-Latent Residual Fusion with the fixed $\hat Z_t$
  \State update the action sample with the inherited solver
\EndFor
\State \Return action chunk $A_t$
\end{algorithmic}
\end{algorithm}

\section{Architecture and Parameter Details}

\begin{table*}[t]
\centering
\caption{Selected tensor and module contracts. $D_\pi$ and $D_{\mathrm{AE}}$
denote the loaded VLM and action-expert widths; $A$ is the benchmark action
dimension. Dimensions not uniquely stored in the aggregate
artifacts are retained symbolically rather than guessed.}
\label{tab:contracts}
\small
\setlength{\tabcolsep}{4pt}
\begin{tabular}{p{0.21\textwidth}p{0.27\textwidth}p{0.23\textwidth}p{0.22\textwidth}}
\toprule
Module & Input & Output & Structure \\
\midrule
Frozen spatial teacher & $B\times3$ RGB views & $B\times768\times2048$ & Frozen VGGT; bilinear token alignment \\
Future Latent Encoder & two $B\times768\times2048$ streams & $B\times8\times256$ & width 256, 2 layers, 8 heads, learned queries \\
Current-Conditioned Decoder & current $B\times768\times2048$, latent $B\times8\times256$ & $B\times768\times2048$ & width 256, 2 cross-attention blocks, zero-init residual output \\
Spatial Grounding Adapter & $B\times N\times D_\pi$ & $S_t:B\times N\times2048$ & LN--MLP; second LN--MLP to zero-gated prefix residual \\
Action Intent Representation & $B\times10\times D_{\mathrm{AE}}$ & $B\times D_{\mathrm{AE}}$ & detached mean over action-token dimension \\
Future Latent Predictor & $S_t$, $B\times D_\pi$ language, $B\times A$ state, $B\times D_{\mathrm{AE}}$ intent & $B\times8\times256$ & width 1024, 3 cross-attention layers, 8 heads \\
Future-Latent Residual Fusion & $B\times10\times D_{\mathrm{AE}}$, $B\times8\times256$ & $B\times10\times D_{\mathrm{AE}}$ & 8-head cross-attention, $2D_{\mathrm{AE}}$ MLP, zero-init gate \\
Action output & $B\times10\times D_{\mathrm{AE}}$ & $B\times10\times A$ & inherited action projection and flow solver \\
\bottomrule
\end{tabular}
\end{table*}

\begin{table}[t]
\centering
\caption{Frozen/trainable boundary for the selected LIBERO-Plus model.}
\label{tab:frozen}
\small
\begin{tabular}{@{}p{0.66\columnwidth}p{0.28\columnwidth}@{}}
\toprule
Component & Status \\
\midrule
VLM Backbone & frozen \\
Action Expert and action projection & frozen \\
Frozen VGGT Spatial Teacher & frozen, training only \\
Future Latent Encoder/Decoder (6.943M) & frozen, training only \\
Spatial Grounding Adapter/injection & trainable \\
Future Latent Predictor/fusion & trainable \\
Other inherited geometry modules & frozen, inactive \\
\midrule
Loaded parameters & 3,720.77M \\
Trainable parameters & 69.94M (1.88\%) \\
Online executed added parameters & 69.94M \\
\bottomrule
\end{tabular}
\end{table}

The loaded SG-LL class has 104.01M more parameters than official
$\pi_{0.5}$, but 34.07M are frozen tokenizer or inherited inactive parameters
and are not online-executed additions.  The measured online additions are
8.40M spatial and 61.53M future-predictor/fusion parameters.

\section{Training and Evaluation Protocols}

\subsection{LIBERO-Plus}

Stage I caches 2,197 usable examples from 1,180 teacher batches (cache batch
2) and trains for 4,000 AdamW steps with batch 8, seed 31, learning rate
$3\times10^{-4}$, weight decay $10^{-4}$, gradient clipping 1.0, and a
1,941/256 train/validation split.  The cache and tokenizer require about
10:44 and 2:44 on one RTX 5090, respectively.

Stage II initializes from official $\pi_{0.5}$-LIBERO and uses the
LIBERO-Plus LeRobot training data.  It freezes the VLM and Action Expert,
uses training seed 42, micro-batch 8, gradient accumulation 2 (effective
batch 16), 10,000 optimizer steps, AdamW with weight decay $10^{-10}$,
gradient clipping 1.0, bfloat16, gradient checkpointing, 5,000 warmup steps,
and peak/end learning rate $5\times10^{-5}$.  One RTX 5090 run takes
18:38:44 and peaks at 22.39/22.86 GB allocated/reserved memory.

The fixed evaluation contains 10,030 episodes: Spatial 2,402, Object 2,518,
Goal 2,591, and LIBERO-10 2,519.  Evaluation seed is 7 with one trial per
manifest entry and no saved videos.  The overall score is episode-weighted,
$\sum_s k_s/\sum_s n_s$, not an equal-suite macro average.  Every formal
paired result requires exact episode keys, zero duplicate keys, zero errors,
and zero unmatched keys.

\subsection{CALVIN D-to-D}

The reported CALVIN experiment is matched low-budget D-to-D adaptation, not
the ABC-to-D leaderboard protocol.  Both arms initialize from the same
official $\pi_{0.5}$ base, use task-D LeRobot data, normalization, state/action
mapping, physical batch 4, training seed 42, 10,000 optimizer steps,
10-action chunks, and train the Action Expert while freezing the VLM.  SG-LL
adds benchmark-specific Stage-I features and its spatial/lookahead modules.
Its Stage-I cache contains 1,650 usable examples; tokenizer training uses
batch 8 and 4,000 steps.  SG-LL/base training takes 4:53:30/1:24:47 and peaks
at 19.33/11.91 GB allocated memory on different RTX 5090 hosts; the ratio is
not treated as a controlled overhead measurement.

Evaluation uses the fixed official 1,000 five-task D-to-D sequence indices
0--999, maximum 360 environment steps per subtask, replanning every five
steps, and no videos.  The metric is mean completed prefix length in each
five-subtask sequence and the rate of completing at least $q\in\{1,\ldots,5\}$
subtasks.  Both methods have exact sequence coverage, zero duplicate indices,
and zero per-sequence errors.

\subsection{Real Robot}

Experiments use the AIRBOT TOK2 platform with two six-degree-of-freedom AIRBOT
Play arms, two-finger grippers, two 720p wrist-mounted RGB cameras, and one
1080p overhead RGB camera.  The official $\pi_{0.5}$ model is fine-tuned with
its standard recipe to form the task-matched baseline; SG-LL uses the same task
demonstrations and optimization-step budget.  Fold clothes and organize desk
use 6.5 and 5.65 hours of demonstrations, batch 32, and 100,000 training
steps; pick up fruit uses 0.26 hours, batch 32, and 20,000 steps.

Training and evaluation sample configurations independently from the same
fixed task-specific distribution.  Object arrangements and spatial placements
change, while the object pool and task definition remain fixed.  The experiment
therefore measures within-distribution generalization to unseen configurations,
not novel-object or out-of-distribution generalization.  Success is judged by
task completion under the archived task-level criterion, and duration is
averaged over successful trials only.

The archived task-level record provides training volume, batch size, update
budget, trial counts, success outcomes, and successful-trial mean duration.
Because trials were not paired and per-trial duration dispersion was not
retained, the real-robot results are reported as task-level descriptive
evidence with Wilson intervals for success rates.

\section{Statistical Analysis}

For two policies $A,B$ evaluated on identical binary episodes, let
$n_{10}$ count $A$-only successes and $n_{01}$ count $B$-only successes.
The two-sided exact McNemar value is
\begin{equation}
p=\min\left(1,2\sum_{i=0}^{\min(n_{10},n_{01})}
{n_{10}+n_{01}\choose i}2^{-(n_{10}+n_{01})}\right).
\end{equation}
For paired success-rate differences, each episode contributes
$\delta_i=y_i^A-y_i^B$.  We draw $N$ indices with replacement 10,000 times
using the archived statistic seed and report the 2.5 and 97.5 percentiles of
$N^{-1}\sum_i\delta_i$.  CALVIN applies the same paired bootstrap to completed
subtask counts per fixed sequence.

For an unpaired descriptive success rate $\hat p=k/n$, the Wilson 95\%
interval with $z=1.959964$ is
\begin{equation}
\frac{\hat p+z^2/(2n)\ \pm\ z
\sqrt{\hat p(1-\hat p)/n+z^2/(4n^2)}}{1+z^2/n}.
\end{equation}
Wilson intervals are not substituted for paired tests.  Training-seed results
are reported separately because repeated checkpoints sharing an evaluation
manifest are not independent episode replicates.

\section{Additional Results}

\subsection{LIBERO-Plus Main and Matched Continued Adaptation}

\begin{table*}[t]
\centering
\caption{Complete LIBERO-Plus evaluation. The matched baseline preserves the
original architecture and receives the same 10,000-step data budget; it trains
the original Action Expert, whereas selected SG-LL freezes it.}
\label{tab:main}
\small
\begin{tabular}{lrrr}
\toprule
Suite & Official $\pi_{0.5}$ & Original architecture, 10k & SG-LL, 10k \\
\midrule
Spatial & 2185/2402 (90.97\%) & 2123/2402 (88.38\%) & 2255/2402 (93.88\%) \\
Object & 2223/2518 (88.28\%) & 2174/2518 (86.34\%) & 2274/2518 (90.31\%) \\
Goal & 2072/2591 (79.97\%) & 1987/2591 (76.69\%) & 2097/2591 (80.93\%) \\
LIBERO-10 & 1990/2519 (79.00\%) & 2000/2519 (79.40\%) & 2021/2519 (80.23\%) \\
\midrule
Aggregate & 8470/10030 (84.45\%) & 8284/10030 (82.59\%) & 8647/10030 (86.21\%) \\
\bottomrule
\end{tabular}
\end{table*}

Against official $\pi_{0.5}$, SG-LL has $n_{10}=661$, $n_{01}=484$,
$n_{11}=7986$, and $n_{00}=899$, for $+1.765$ pp, exact
$p=1.876\times10^{-7}$, and bootstrap 95\% CI $[+1.147,+2.423]$ pp
(1,000 archived resamples).  Against the matched original-architecture 10k
baseline, the counts are $901/538/7746/845$ in the same order, giving
$+3.619$ pp, $p=8.4903\times10^{-22}$, and 10,000-sample CI
$[+2.901,+4.377]$ pp.  Per-suite SG-LL gains over that baseline are
$+5.496/+3.971/+4.245/+0.834$ pp.  Spatial, Object, and Goal are significant;
LIBERO-10 is not ($p=0.37254$).  This comparison matches initialization,
architecture baseline, dataset, and optimizer-step budget, but not the
trainable-parameter set.

\begin{table*}[t]
\centering
\caption{Per-suite paired statistics for SG-LL against the official checkpoint
and the original-architecture 10k continued-adaptation baseline. $n_{10}$ is an
SG-LL-only success and $n_{01}$ is a baseline-only success.}
\label{tab:paired-main-suite}
\scriptsize
\setlength{\tabcolsep}{4pt}
\begin{tabular}{llrrrr}
\toprule
Reference & Suite & $n_{10}$ & $n_{01}$ & Difference (pp) & Exact $p$ \\
\midrule
Official & Spatial & 129 & 59 & +2.91 & $3.59\!\times\!10^{-7}$ \\
Official & Object & 137 & 86 & +2.03 & $7.76\!\times\!10^{-4}$ \\
Official & Goal & 146 & 121 & +0.96 & $1.42\!\times\!10^{-1}$ \\
Official & LIBERO-10 & 249 & 218 & +1.23 & $1.65\!\times\!10^{-1}$ \\
\midrule
Continued 10k & Spatial & 207 & 75 & +5.50 & $1.85\!\times\!10^{-15}$ \\
Continued 10k & Object & 187 & 87 & +3.97 & $1.47\!\times\!10^{-9}$ \\
Continued 10k & Goal & 245 & 135 & +4.25 & $1.80\!\times\!10^{-8}$ \\
Continued 10k & LIBERO-10 & 262 & 241 & +0.83 & $3.73\!\times\!10^{-1}$ \\
\bottomrule
\end{tabular}
\end{table*}

\subsection{Training-Seed Checks}

\begin{table}[t]
\centering
\caption{SG-LL Spatial training-seed checks against official $\pi_{0.5}$ on
the same 2,402-episode manifest.}
\small
\begin{tabular}{crrr}
\toprule
Seed & Success & Gain (pp) & Exact $p$ \\
\midrule
42 & 2255/2402 & +2.91 & $<0.001$ \\
43 & 2229/2402 & +1.83 & 0.000919 \\
44 & 2213/2402 & +1.17 & 0.035799 \\
\bottomrule
\end{tabular}
\end{table}

For Object, Goal, and LIBERO-10, additional seeds 43/44/45 obtain
90.51/89.56/90.31\%, 81.51/81.09/80.97\%, and
83.76/83.25/83.01\%, respectively.  Object and LIBERO-10 are positive for
every listed seed; Goal is positive for all three but only seed 43 is
conventionally significant.  No pooled seed-level McNemar test is formed.

\subsection{CALVIN Full-1,000 Results}

\begin{table}[t]
\centering
\caption{Matched low-budget CALVIN D-to-D evaluation.}
\small
\begin{tabular}{lrrr}
\toprule
Metric & Base 10k & SG-LL 10k & Difference \\
\midrule
Mean completed & 0.580 & 0.725 & +0.145 \\
$\geq1$ & 403/1000 & 442/1000 & +3.9 pp \\
$\geq2$ & 129/1000 & 182/1000 & +5.3 pp \\
$\geq3$ & 36/1000 & 64/1000 & +2.8 pp \\
$\geq4$ & 11/1000 & 26/1000 & +1.5 pp \\
$\geq5$ & 1/1000 & 11/1000 & +1.0 pp \\
\bottomrule
\end{tabular}
\end{table}

The paired bootstrap 95\% CI for the mean difference is
$[+0.075,+0.217]$.  SG-LL-only/base-only counts for thresholds one through
five are $182/143$, $130/77$, $53/25$, $25/10$, and $11/1$; exact paired
$p$-values are 0.03488, 0.0002810, 0.002031, 0.01667, and 0.006348.

\subsection{Target Semantics and Token Capacity}

With the teacher frozen, token probes trained from future-only versus
current-only features obtain validation losses 0.011462 and 0.297392.
Future-decoding probes obtain 0.074566 with current+token, 0.093445 with
current only, and 0.124070 with token only.  The frozen decoder itself obtains
0.076249 with the correct current reference and 0.789774 with a zero current
reference.  Together, these results support a future-dominant,
current-complementary current-conditioned future latent.

\begin{table}[t]
\centering
\caption{Matched token-count semantic audit. Lower probe loss is better.}
\footnotesize
\setlength{\tabcolsep}{3pt}
\begin{tabular}{crrr}
\toprule
Tokens & Future probe & Current probe & Joint decode \\
\midrule
1 & 0.01656 & 0.33460 & 0.07580 \\
2 & 0.03899 & 0.22233 & 0.08162 \\
4 & 0.02951 & 0.19692 & 0.07983 \\
8 & 0.01146 & 0.29739 & 0.07457 \\
16 & 0.02950 & 0.22864 & 0.07713 \\
\bottomrule
\end{tabular}
\end{table}

Across $M\in\{1,2,4,8,16\}$, future-only prediction remains substantially
easier than current-only prediction, while joint decoding remains strongest.
The same future-dominant, current-complementary pattern across bottleneck sizes
shows that the observed representation is not an artifact of the selected
eight-token setting.

\subsection{Future-Frame and Fusion-Disabled Controls}

The architecture-matched future-frame latent control removes the
sample-specific current target from Stage-I target construction while keeping
eight 256-dimensional tokens, student predictor, fusion, training budget, and
evaluation protocol matched.  It obtains 8588/10030 (85.623\%), compared with
8470/10030 (84.447\%) for official $\pi_{0.5}$ and 8647/10030 (86.211\%) for
SG-LL.  SG-LL versus frame-token has contingency
$n_{10}/n_{01}/n_{11}/n_{00}=509/450/8138/933$, difference $+0.588$ pp,
$p=0.061023$, and CI $[-0.010,+1.186]$ pp.  Direct future-frame prediction is
therefore already a strong design, while current-conditioned abstraction gives
the best numerical aggregate with a modest additional gain.

The separately trained fusion-disabled control retains spatial grounding but
disables Future-Latent Residual Fusion throughout training and evaluation.  It
scores 2215/2402, 2276/2518, 2115/2591, and 2095/2519, aggregating to
8701/10030 (86.75\%).  SG-LL versus this control has
$469/523/8178/860$, difference $-0.538$ pp, $p=0.092373$, and CI
$[-1.157,+0.080]$ pp.  The effect is suite-dependent: Spatial significantly
favors SG-LL ($+1.665$ pp), Object is effectively neutral, and Goal and
LIBERO-10 favor the control.  This complements the fixed-policy interventions
below by separating behavioral reliance from the contribution of an
independently trained architectural pathway.

\begin{table}[!htbp]
\centering
\caption{Per-suite paired statistics for the future-frame and trained
fusion-disabled controls. Differences are SG-LL minus the listed control.}
\label{tab:control-suite-stats}
\scriptsize
\setlength{\tabcolsep}{2.3pt}
\begin{tabular}{lrrrr}
\toprule
Suite & SG only & Ctrl only & $\Delta$ (pp) & Exact $p$ \\
\midrule
\multicolumn{5}{l}{\textit{Future-frame control}} \\
Spatial & 87 & 63 & +1.00 & $6.00\!\times\!10^{-2}$ \\
Object & 92 & 73 & +0.75 & $1.61\!\times\!10^{-1}$ \\
Goal & 105 & 117 & -0.46 & $4.60\!\times\!10^{-1}$ \\
LIBERO-10 & 225 & 197 & +1.11 & $1.89\!\times\!10^{-1}$ \\
\midrule
\multicolumn{5}{l}{\textit{Fusion-disabled control}} \\
Spatial & 98 & 58 & +1.67 & $1.70\!\times\!10^{-3}$ \\
Object & 82 & 84 & -0.08 & $9.38\!\times\!10^{-1}$ \\
Goal & 109 & 127 & -0.69 & $2.68\!\times\!10^{-1}$ \\
LIBERO-10 & 180 & 254 & -2.94 & $4.45\!\times\!10^{-4}$ \\
\bottomrule
\end{tabular}
\end{table}
\FloatBarrier

\subsection{Fixed-Policy Mechanism Interventions}

For the same seed-42 checkpoint and exact 2,402 Spatial episodes, normal
episode-matched latent content gives 2255 successes.  Random, norm-matched
random, and cross-episode content give 2170, 2167, and 2218 successes; normal
minus intervention differences are $+3.539$, $+3.664$, and $+1.540$ pp with
$p=2.65\times10^{-9}$, $2.78\times10^{-9}$, and 0.00416.  Zeroing the
Action Intent Representation gives 2213 successes, a $+1.748$ pp normal-minus-
zero difference ($p=0.001207$, CI $[+0.708,+2.748]$ pp).  These interventions show that the deployed policy functionally uses
both episode-matched latent content and Action-Expert-derived context.  They
measure fixed-policy behavioral dependence and are complementary to the
separately trained controls above.

An exact fusion-scale intervention gives 2175/2402 at $\alpha=0$ versus
2255/2402 at $\alpha=1$ on one execution site ($p=5.9982\times10^{-9}$,
normal-minus-zero CI $[+2.206,+4.455]$ pp).  An independent-site replication
gives 2195/2402 at $\alpha=0$ ($p=8.1096\times10^{-6}$, CI
$[+1.415,+3.580]$ pp).  This provides replicated fixed-policy evidence that the trained
Spatial policy uses a nonzero fusion residual.

\subsection{Residual Diagnostics}

The completed diagnostic archive contains 4,921 episodes and 212,896 policy
requests over Spatial and LIBERO-10.  For each episode we summarize the hidden
fusion-residual norm, residual-to-base ratio, projected velocity-correction
norm, and correction-to-base ratio over all requests and over the final five
and ten environment steps.  All 24 failure-minus-success comparisons are
positive, and all 10,000-sample bootstrap intervals exclude zero.

\begin{table}[!htbp]
\centering
\caption{Representative failure-minus-success residual differences. Values in
brackets are paired-bootstrap 95\% intervals.}
\label{tab:residual-diagnostics}
\scriptsize
\setlength{\tabcolsep}{2.5pt}
\begin{tabular}{lcc}
\toprule
Suite/window & Hidden residual $\ell_2$ & Velocity correction $\ell_2$ \\
\midrule
Spatial/all & 0.1257 [0.1094, 0.1428] & 0.0739 [0.0645, 0.0838] \\
Spatial/final 5 & 0.1507 [0.1164, 0.1869] & 0.0904 [0.0703, 0.1116] \\
LIBERO-10/all & 0.0590 [0.0504, 0.0679] & 0.0353 [0.0303, 0.0405] \\
LIBERO-10/final 5 & 0.0952 [0.0766, 0.1147] & 0.0572 [0.0463, 0.0686] \\
\bottomrule
\end{tabular}
\end{table}
\FloatBarrier

Larger residual corrections are observed more often in failed episodes,
especially near termination.  This outcome-conditioned diagnostic motivates
future work on adaptive gating and broader temporal scope; it is reported as a
correlational pattern rather than a serving-policy intervention.

\subsection{Additional Scope Checks}

Two completed checks further characterize the selected design.  First,
unfreezing only the final two Action Expert Transformer blocks at 0.1 times the
planning learning rate gives 2241/2402 for SG-LL and 2232/2402 for its matched
fusion-disabled control ($+0.375$ pp, $p=0.517921$); normal versus zero-latent
serving gives 2241/2402 versus 2224/2402 ($+0.708$ pp, $p=0.20718$).  Thus the
frozen-Action-Expert configuration remains the clearest setting for isolating
the added SG-LL pathway.

Second, an isolated standard-LIBERO Spatial check operates near ceiling:
official $\pi_{0.5}$ obtains 492/500 (98.4\%), SG-LL obtains 490/500 (98.0\%)
and 492/500 (98.4\%) across two evaluations, and the spatial/fusion-disabled
control obtains 487/500 (97.4\%).  We therefore use LIBERO-Plus as the primary
robustness benchmark, where the fixed manifest offers substantially more room
to resolve policy differences.

\subsection{Real-Robot Results}

\begin{table}[!htbp]
\centering
\caption{Descriptive real-robot results. Wilson intervals quantify uncertainty
for each unpaired success rate. Duration is averaged over successful trials;
raw duration dispersion is unavailable.}
\label{tab:real-robot-supp}
\scriptsize
\setlength{\tabcolsep}{3pt}
\begin{tabular}{llrrr}
\toprule
Task & Method & Success & Wilson 95\% CI & Mean time \\
\midrule
Fold & Base & 9/20 (45\%) & [25.82, 65.79]\% & 120 s \\
Fold & SG-LL & 12/20 (60\%) & [38.66, 78.12]\% & 106 s \\
Fruit & Base & 10/10 (100\%) & [72.25, 100.00]\% & 34 s \\
Fruit & SG-LL & 10/10 (100\%) & [72.25, 100.00]\% & 31 s \\
Desk & Base & 13/20 (65\%) & [43.29, 81.88]\% & 138 s \\
Desk & SG-LL & 15/20 (75\%) & [53.13, 88.81]\% & 115 s \\
\bottomrule
\end{tabular}
\end{table}
\FloatBarrier

Observed success is higher on two tasks and tied on one, while
successful-trial mean duration is lower on all three.  Because trials are not
paired and per-trial duration dispersion is unavailable, we present these
results as task-level descriptive evidence rather than aggregate significance
tests.

\section{Efficiency and Reproducibility Notes}

On one RTX 5090 with synthetic batch-one input and ten flow steps, official
$\pi_{0.5}$ and SG-LL require $292.11\pm1.63$ and
$370.27\pm1.85$ ms per action-chunk request.  Peak allocated memory is
7.647 and 8.156 GB.  SG-LL therefore adds 26.76\% model-only latency and
0.510 GB (6.665\%) peak memory in this controlled measurement.  These are not
simulator FPS, network latency, or real-robot control-frequency measurements.

The SG-LL code, including the model implementation on $\pi_{0.5}$, training
configurations and entry points, and evaluation scripts, is available at
\url{https://github.com/Cqhddyb/SG-LL}.  Model weights, datasets, caches, and
training artifacts are not included in the repository.
Only completed, archived studies are included in the reported results.

\end{document}